\documentclass{article}

\usepackage[preprint]{neurips_2023}

\usepackage[utf8]{inputenc}
\usepackage[T1]{fontenc}
\usepackage{hyperref}
\usepackage{url}
\usepackage{booktabs}
\usepackage{amsfonts}
\usepackage{amsmath}
\usepackage{amssymb}
\usepackage{amsthm}
\usepackage{mathtools}
\usepackage{nicefrac}
\usepackage{microtype}
\usepackage{xcolor}
\usepackage{algorithm}
\usepackage{algpseudocode}
\usepackage{tikz}
\usetikzlibrary{positioning,fit,arrows.meta,calc}
\tikzset{
  io/.style   ={draw=blue!50!black, fill=blue!10,   rounded corners=2pt, align=center, inner sep=3pt, font=\footnotesize},
  proc/.style ={draw=black!55,      fill=black!4,   rounded corners=2pt, align=center, inner sep=3pt, font=\footnotesize},
  procf/.style={draw=green!45!black, fill=green!11,  rounded corners=2pt, align=center, inner sep=3pt, font=\footnotesize},
  procs/.style={draw=orange!78!black, fill=orange!14, rounded corners=2pt, align=center, inner sep=3pt, font=\footnotesize},
  procv/.style={draw=violet!70!black, fill=violet!11, rounded corners=2pt, align=center, inner sep=3pt, font=\footnotesize},
  gate/.style ={draw=red!55!black,  fill=orange!22, rounded corners=2pt, align=center, inner sep=3pt, font=\footnotesize, thick},
  accept/.style={draw=green!50!black, fill=green!18, rounded corners=2pt, align=center, inner sep=3pt, font=\footnotesize},
  hold/.style ={draw=red!60!black,  fill=red!12,    rounded corners=2pt, align=center, inner sep=3pt, font=\footnotesize},
  store/.style={draw=black!60,      fill=yellow!20, rounded corners=2pt, align=center, inner sep=2.5pt, font=\scriptsize},
  cert/.style ={draw=black!72,      fill=yellow!26, rounded corners=2pt, align=center, inner sep=3pt, font=\footnotesize},
  arr/.style  ={-{Stealth[length=2mm]}, semithick},
  flab/.style ={font=\scriptsize, fill=white, inner sep=1pt},
}
\usepackage{enumitem}
\usepackage{fancyhdr}

\fancypagestyle{plain}{%
  \fancyhf{}%
  \fancyfoot[C]{\thepage\\[2pt]\footnotesize 2026 JP Morgan Chase \& Co. All rights reserved}%
}

\makeatletter
\renewcommand{\@noticestring}{%
  Preprint. Under review.\\
  2026 JP Morgan Chase \& Co. All rights reserved%
}
\makeatother

\newcommand{\KL}{\mathrm{KL}}
\newcommand{\Wone}{W_1}
\newcommand{\E}{\mathbb{E}}
\newcommand{\R}{\mathbb{R}}
\newcommand{\dist}{\mathcal{D}}
\newcommand{\pol}{\pi}
\newcommand{\NSF}{\textsc{nsf}}
\newcommand{\alg}[1]{\textsc{Alg~#1}}
\newcommand{\code}[1]{\texttt{#1}}

\title{Self-Evolving Agents with Anytime-Valid Certificates}

\author{%
  Biswa Sengupta\\
  LLM Suite Team, JPMorgan Chase \& Co.\\
  \texttt{biswa.sengupta@jpmorgan.com}\\
}

\begin{document}

\maketitle

\begin{abstract}
Self-evolving agents violate the assumption behind most learning-theoretic guarantees: the data, evaluator, components, and hypothesis space are produced by the policy being updated. We present \textbf{SEA}, an architecture that confines self-modification to a small steering adapter and a versioned harness around a \emph{frozen} base model and admits each modification only through an anytime-valid gate that emits an auditable certificate against a fixed error budget. Five loop controllers compose published guarantees; because such gates can only \emph{select} among behaviors the frozen base already produces, five verifier-in-the-loop mechanisms---best-of-$N$, micro-step search, self-authored reproduction oracles, search-layer control, and self-repair---supply the dense, grader-free signal the gates require, computed from the issue text alone. On a $52$-instance SWE-bench Verified subset across four base models, base capability is the dominant, confound-free effect, and on two strong base models a deliberate no-op-composite control isolates the suite's contribution at $+4$ and $+5$ (\textsc{Glm}~5.2 $24\to28$; \textsc{Gpt} $29\to34$, the $65\%$ best), with event logs confirming that its mechanisms fire and prevent regressions. Results are single-run on expensive evaluations; confirming run-to-run variance and adapting the per-task algorithm mix are future work.
\end{abstract}

\begin{center}
\begin{minipage}{0.92\textwidth}
\footnotesize
\textbf{Disclaimer:} This paper was prepared for informational purposes by the LLM Suite group of JP Morgan Chase and its affiliates (`JPMC') and is not a product of the Research Department of JP Morgan. JP Morgan makes no representation, warranty or undertaking whatsoever and disclaims all liability for the completeness, accuracy or reliability of the information contained herein. This document is not intended as investment research or investment advice, or a recommendation, offer or solicitation for the purchase or sale of any security, financial instrument, financial product or service, or to be used in any way for evaluating the merits of participating in any transaction, and shall not constitute a solicitation under any jurisdiction or to any person, if such solicitation under such jurisdiction or to such person would be unlawful.
\end{minipage}
\end{center}

\section{Introduction}
\label{sec:intro}

A self-evolving agent improves its own future behavior using data, evaluations, components, and a hypothesis space that it itself produces---rewriting prompts and tools, distilling its outputs, learning its reward models, growing skill libraries. The guarantees one would invoke for such systems, however, were proven for \emph{exogenous} environments: continual-learning forgetting bounds~\citep{farajtabar2020orthogonal,chugg2023unified}, convergence of preference optimization~\citep{tiapkin2025nashmp,wang2025magnetic}, unbiasedness of policy-gradient estimators~\citep{meulemans2023cocoa}, safe policy improvement~\citep{thomas2015high}, and library-learning optimality~\citep{bowers2023stitch} each assume a task stream, evaluator, MDP, or program library fixed independently of the learner. We call the violation the \emph{endogenous-loop failure mode}: the evolving policy generates the data it trains on, the evaluator it is judged by, the components it is built from, and the hypothesis space it searches. The name is a shorthand, not a precise mathematical category, and taken literally it overstates the problem: violating a theorem's hypotheses voids its certificate but does not negate its conclusion. The guarantee simply ceases to be \emph{certified}---the bound may still hold, may degrade gracefully, or may break, depending on the problem---and performative-prediction theory shows the loop can in fact still contract when the policy-induced distribution shift is small enough~\citep{perdomo2020performative}. We use the term to mark where classical guarantees stop applying, not to claim that learning provably fails.

This paper develops an architecture and a set of concrete algorithms for this setting, together with an executable reference implementation from which all pseudo-code in this paper is distilled. Two principles organize the design. First, every self-modification passes through an \emph{anytime-valid gate}: each classical seed result is wrapped in exactly the machinery required for its guarantee to survive the closed loop---performative stability~\citep{perdomo2020performative}, anytime-valid inference~\citep{ramdas2023game,howard2021time}, dynamic-regret online learning~\citep{cutkosky2020parameter,baby2022optimal}, and two-timescale stochastic approximation~\citep{borkar2008stochastic}. Second, gates can only \emph{select} among behaviors the frozen base model already produces; when a base model's failure is systematic rather than stochastic, and the reward arrives only at the end of an episode, there is nothing for a gate to select. We therefore make the task verifier an active in-loop control signal (\S\ref{sec:verifier})---inside the episode, across attempts, over the action space, and inside credit assignment---so that the controllers have both the signal and the variation they need to act on.

\begin{figure}[t]
\centering
\begin{tikzpicture}[font=\footnotesize,
  layer/.style={draw=blue!45!black!60, rounded corners=2pt, minimum width=5.2cm, minimum height=6.6mm, align=center, fill=blue!8},
  slowc/.style={draw=orange!75!black, rounded corners=2pt, text width=5.7cm, inner ysep=2.6pt, align=left, fill=orange!14},
  fastc/.style={draw=green!45!black, rounded corners=2pt, text width=5.7cm, inner ysep=2.6pt, align=left, fill=green!12},
  guardc/.style={draw=violet!70!black, rounded corners=2pt, text width=5.7cm, inner ysep=2.6pt, align=left, fill=violet!10},
  aux/.style={draw=black!70, rounded corners=2pt, align=center, fill=yellow!16, inner ysep=3pt},
  arr/.style={-{Stealth[length=2.2mm]}, semithick},
  lab/.style={font=\scriptsize, fill=white, inner sep=1pt}
]
\node[layer, fill=black!14, draw=black!60] (L0) {$L_0$ --- frozen base model};
\node[layer, above=3.2mm of L0] (L1) {$L_1$ --- steering adapter $\theta$ (online-steered; not fine-tuned)};
\node[layer, above=3.2mm of L1] (L2) {$L_2$ --- harness: prompt $\cdot$ tools $\cdot$ budgets $\cdot$ library};
\node[draw=blue!45!black!70, dashed, rounded corners=3pt, fit=(L0)(L1)(L2), inner sep=2.6mm,
      label={[font=\scriptsize, text=blue!45!black]90:{deployed policy $\pol_t=L_0\circ L_1^{(t)}\circ L_2^{(t)}$}}] (POL) {};
\node[slowc, anchor=north west] (A4) at ($(L2.north east)+(22mm,4mm)$)
  {\alg{4} SGM-CS: \scriptsize gated harness edits (CS $+$ $\varepsilon\Wone$ $+$ CTHS)};
\node[slowc, below=2.4mm of A4] (A5) {\alg{5} SDC-QD: \scriptsize grows the abstraction library (MDL $+$ QD)};
\node[fastc, below=2.4mm of A5] (A3) {\alg{3} PA-COCOA: \scriptsize counterfactual credit $\to\theta$};
\node[fastc, below=2.4mm of A3] (A1) {\alg{1} PPB-CL: \scriptsize forgetting gate $+$ trust region on $\theta$};
\node[guardc, below=2.4mm of A1] (A2) {\alg{2} PNMP-A: \scriptsize anchors the learned reward model $q$};
\node[draw=black!70, dashed, rounded corners=3pt, fit=(A4)(A5)(A3)(A1)(A2), inner sep=2.4mm,
      label={[font=\scriptsize]90:{$L_3$ --- loop controllers (composite two-timescale scheduler)}}] (CTRL) {};
\draw[arr, orange!75!black] (A4.west) to[out=180,in=0] node[lab,pos=.38,above=-.5pt]{edit (slow)} ([yshift=2.2mm]L2.east);
\draw[arr, orange!75!black] (A5.west) to[out=180,in=0] node[lab,pos=.62,below=-.5pt]{grow (slow)} ([yshift=-2.2mm]L2.east);
\draw[arr, green!45!black] (A3.west) to[out=180,in=0] node[lab,pos=.38,above=-.5pt]{train (fast)} ([yshift=2.2mm]L1.east);
\draw[arr, green!45!black] (A1.west) to[out=180,in=0] node[lab,pos=.66,below=-.5pt]{protect (fast)} ([yshift=-2.2mm]L1.east);
\node[aux, minimum width=5.2cm] (ENV) at ($(L0 |- CTRL.south)+(0,-17mm)$)
  {environment $+$ verifier\\[-2pt]{\scriptsize tasks $\cdot$ test execution $\cdot$ rewards $R,\tilde R\in[0,1]$}};
\node[aux, fill=violet!10, draw=violet!70!black] (RM) at ($(CTRL.west |- ENV)+(11mm,0)$) {reward\\[-2pt] model $q$};
\node[aux, fill=black!8, draw=black!70] (LED) at ($(CTRL.east |- ENV)+(-16mm,0)$)
  {certificate ledger\\[-2pt]{\scriptsize \textsc{accept}/\textsc{hold}/\NSF{} $\cdot$ $\delta$-spend}};
\draw[arr] ([xshift=-13mm]POL.south) -- node[lab, left=2pt]{deploy / act} ([xshift=-13mm]ENV.north);
\draw[arr] ([xshift=13mm]ENV.north) -- node[lab, right=2pt]{obs.\ $+$ test feedback} ([xshift=13mm]POL.south);
\draw[arr, blue!55!black] ([xshift=25mm]ENV.north) to[out=62,in=180] node[lab, pos=.62, above=2pt]{rollouts $\cdot$ rewards}
      ($(CTRL.west)+(0,-21.5mm)$);
\draw[arr, violet!70!black] ([xshift=-17mm]CTRL.south) -- node[lab, pos=.72, right=2pt]{e-value drift gate} (RM.north);
\draw[arr, violet!70!black, dashed] (RM.west) -- node[lab, pos=.5, below=1pt]{\scriptsize scores rollouts} (ENV.east |- RM.west);
\draw[arr, black!70] ([xshift=16mm]CTRL.south) -- (LED.north);
\end{tikzpicture}
\caption{How the four layers and the five controllers interact. The deployed policy $\pol_t=L_0\circ L_1^{(t)}\circ L_2^{(t)}$ composes a frozen base model $L_0$, a small steering adapter $L_1$, and a mutable harness $L_2$; the fourth layer---the $L_3$ loop controllers---sits \emph{outside} this forward pass. Deployments induce the performative distribution $\dist(\pol_t)$ and the verifier returns bounded terminal and process rewards; the $L_3$ controllers consume these and act on their own layers through anytime-valid gates---\alg{3}/\alg{1} update and protect $L_1$ each round (fast timescale), \alg{4}/\alg{5} edit and grow $L_2$ every $K$ rounds (slow timescale), and \alg{2} guards any learned reward model with an e-value drift gate. Every decision is logged in the certificate ledger against the error budget $\delta_0$. In all reported runs $L_1$ is \emph{steered}---its distribution over discrete directives updated online by \alg{1}/\alg{3}---never weight-fine-tuned; weight-level adapter training by the slow-loop distillation of \S\ref{sec:twoloop} is designed but not run here.}
\label{fig:arch}
\end{figure}
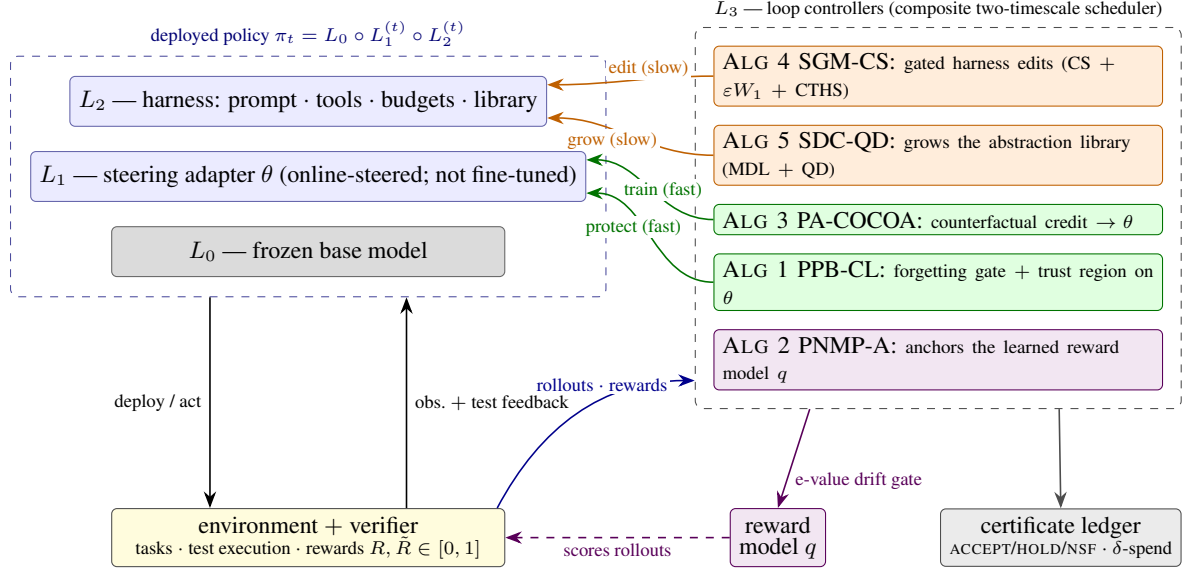

\paragraph{Contributions.}
\begin{enumerate}[leftmargin=1.5em,itemsep=1pt,topsep=2pt]
\item \textbf{A four-layer reference architecture} (\S\ref{sec:architecture}) that decomposes a self-evolving LLM agent into a frozen base model $L_0$, a small steering adapter $L_1$ (steered online; not weight-fine-tuned in any reported run), a mutable, versioned harness $L_2$, and a loop controller $L_3$ (Figure~\ref{fig:arch}). Because $L_0$ is frozen and $L_1$ is low-dimensional, policy deltas $\lVert\pol_t - \pol_{t-1}\rVert$ are measurable and can be trust-regioned, which is what renders the performative-sensitivity machinery applicable at all.
\item \textbf{Five loop controllers} (\S\ref{sec:algorithms}), one per failure mode of the endogenous loop: stability--plasticity, self-referential collapse, credit assignment, verifiable self-modification, and hypothesis-space expansion. Each is given as a precise problem statement, an algorithmic solution with pseudo-code (Algorithms~\ref{alg:ppbcl}--\ref{alg:sdcqd}), and an explicit account of which published results it builds on and which guarantees remain open conjectures in the endogenous setting.
\item \textbf{Verifier-in-the-loop mechanisms and a two-loop design} (\S\ref{sec:verifier}): a graded verifier (Eq.~\ref{eq:graded}), closed-loop test execution, an explore$\to$edit budget, process-level reward, and a verifier-gated refinement hill-climb (Alg.~\ref{alg:verifier}), building to a verified micro-step search (Alg.~\ref{alg:micro}) and a search-then-distill two-loop design~\citep{zelikman2022star,gulcehre2023reinforced} into which the five controllers are re-aimed (Table~\ref{tab:reaim}; four on the search layer, Alg.~\ref{alg:continual}, validated by offline gate simulations).
\item \textbf{A self-authored in-loop verifier} (\S\ref{sec:firewall}, Algorithm~\ref{alg:firewall}): the search is steered by a reproduction-oracle suite the model writes from the issue alone, admitted by a single rule (an oracle must \emph{fail on the unpatched base}), while the held-out grader is reserved for terminal measurement. Where an oracle suite is admitted the search runs grader-free, and the held-out tests never steer it.
\item \textbf{Verified self-repair of the harness} (\S\ref{sec:selfrepair}, Algorithm~\ref{alg:repair}): a repertoire of harness-repair primitives that the loop \emph{selects by measured fix-rate against the real environment}, not by human judgment---the same propose-and-gate discipline as \alg{4}, applied to the agent's own failure modes.
\item \textbf{A reusable anytime-valid statistical core} (\S\ref{sec:stats}): normal-mixture confidence sequences, Hoeffding e-processes with predictable plug-in betting, a \emph{horizon-free, normalized} confirm-triggered harmonic spending schedule, time-uniform PAC-Bayes penalties, parameter-free coin-betting oracles with drift-triggered restarts, exact 1-D Wasserstein computation, wild-bootstrap trend tests, MAP-Elites archives, and Stitch-style MDL compression by antiunification with sound dominance pruning.
\end{enumerate}

We are deliberate about epistemic status. Every mathematical statement in this paper is drawn from a published result, cited at the point of use; we do not derive new bounds or rates here. Where a classical guarantee is invoked, we state precisely what its source proves and treat its survival under the endogenous loop as an \emph{open conjecture}, without positing a proof. Likewise, whether any given mechanism lifts a given base model is not argued from anecdote: the protocol of \S\ref{sec:experiments} is designed to answer it, and we defer that discussion to the results.

\section{Related Work}
\label{sec:related}

SEA introduces no new learning theory; it reuses guarantees proved for \emph{exogenous} settings and asks what each needs in order to survive a loop the learner closes on itself. The prior work accordingly plays two roles. A shared lens (performativity) and a shared gate (anytime-valid inference) cut across all five controllers; four classical literatures then each supply the seed result for one controller, and a fifth thread---verifier-guided search---is the engine that feeds the gates rather than a controller in its own right.

\paragraph{Performativity is the lens.} The common difficulty is that an agent's own updates move the distribution it is then judged on. Performative prediction makes this precise: \citet{perdomo2020performative} define decision-dependent distributions $\dist(\pol)$ through the sensitivity condition $\Wone(\dist(\pol),\dist(\pol'))\le\varepsilon\lVert\pol-\pol'\rVert$ (their Def.~3.1) and show that for $\beta$-smooth, $\gamma$-strongly-convex losses repeated retraining contracts to a performatively stable point at a linear rate when $\varepsilon<\gamma/\beta$ (their Thm.~3.5), a threshold shown tight in their Prop.~3.6. \citet{mandal2023performative} carry this into reinforcement learning, where reward and transition kernels are $(\varepsilon_r,\varepsilon_p)$-sensitive in the occupancy measure and convergence is \emph{restored by} sufficiently strong regularization (their Thm.~1)---so under performativity, trust regions and regularization are necessary rather than optional. We adopt this lens throughout: the deployed configuration is our performative variable, and $\varepsilon$ enters both as a trust-region coefficient (\alg{1}) and as a confidence-radius inflation (\alg{4}).

\paragraph{Anytime-valid inference is the gate.} Because a self-evolving agent inspects its own statistics every round, any test it relies on must stay valid under continuous peeking---fixed-$n$ inference is invalid by construction. E-values, e-processes, and time-uniform confidence sequences supply exactly this license for optional stopping~\citep{ramdas2023game,howard2021time,ville1939etude}. The Statistical G\"odel Machine~\citep{wu2025sgm}, a statistical descendant of Schmidhuber's proof-based G\"odel machine~\citep{schmidhuber2003godel}, applies the idea to self-modification: it gates each self-edit with an e-value and, because an accepted edit is an irreversible commit, controls familywise (not false-discovery) error under a harmonic spending schedule. \alg{4} builds directly on this gate, adding a performative correction and a non-stationarity test~\citep{chandak2020safe} and inheriting the abstention semantics of high-confidence, Seldonian policy improvement~\citep{thomas2015high,thomas2019preventing}---the source of our ``no solution found'' output.

\paragraph{Four endogenous failure modes, four seeds.} Each remaining controller takes a guarantee proved in an exogenous world and asks it to hold once the agent supplies its own data. \emph{Continual learning.} Orthogonal gradient descent leaves earlier-task predictions unchanged~\citep{farajtabar2020orthogonal}, but its no-forgetting guarantee is an NTK-regime result that needs unbounded Jacobian memory~\citep[Thm.~2]{bennani2020generalisation}, while the time-uniform PAC-Bayes forgetting certificates of \citet[Thm.~3.1]{friedman2025forgetting} and \citet[Thm.~3.1]{chugg2023unified} require a data-independent prior; both assumptions break when the agent reuses its own evolving policy on a stream it generates, which \alg{1} confronts with a finite direction buffer inside a performative trust region. \emph{Self-consuming loops and preference games.} Pure self-training collapses---tails vanish and variance shrinks~\citep{shumailov2024collapse}---unless real data is retained: accumulating rather than replacing data bounds the error independently of iteration count~\citep{gerstgrasser2024accumulate}, and a constant real-data fraction $\alpha$ caps the cumulative shift at $2M(1-(1-\alpha)^t)\alpha^{-1}d_{\mathrm{TV}}(n)$~\citep[Thm.~1]{fu2025collapse}. In parallel, preference optimization converges only in the last iterate and only for a \emph{fixed} game---magnetic mirror descent reaches the regularized Nash equilibrium and recovers the true one by refreshing its magnet~\citep{wang2025magnetic}, and proximal-point self-play contracts geometrically up to a gradient-residual floor~\citep{tiapkin2025nashmp}---while a learned reward model over-optimizes, its proxy--gold gap growing with $\sqrt{\KL}$~\citep{gao2023scaling}. \alg{2} couples a real-data anchor against collapse with last-iterate self-play, separated onto two timescales~\citep{borkar2008stochastic}. \emph{Credit assignment.} Hindsight credit assignment reweights each reward by how much the action could have influenced it, dropping rewards it could not~\citep{harutyunyan2019hindsight}; COCOA generalizes the conditioning from future states to rewarding outcomes and is unbiased under a fully-predictive encoding with ground-truth contribution coefficients (their Def.~2 and Thm.~1)---but only in a \emph{fixed} MDP~\citep{meulemans2023cocoa}. \alg{3} folds the deploying policy into the outcome encoding and drives the update with parameter-free, strongly-adaptive online learning~\citep{cutkosky2020parameter,orabona2016coin,baby2022optimal}, whose dynamic-regret rates are stated against a drifting comparator. \emph{Library learning.} DreamCoder grows a program library by wake-sleep MDL learning~\citep{ellis2021dreamcoder} and Stitch makes the abstraction step exact via corpus-guided top-down synthesis with sound dominance pruning~\citep{bowers2023stitch}, but neither has a convergence or no-collapse guarantee when the corpus is generated under the very library being learned; \alg{5} therefore keeps a MAP-Elites archive over behaviors~\citep{mouret2015illuminating,cully2018quality} rather than a single library, and certifies held-out description length in the spirit of PAC-Bayes lifelong learning~\citep{pentina2015pacbayesian}.

\paragraph{Verifier-guided search is the engine.} Because the controllers can only \emph{select} among behaviors the frozen base already produces, a separate line of work supplies the variation and the dense signal they act on. Coverage under repeated sampling grows along an approximate exponentiated power law, yet only an automatic verifier turns that coverage into solved instances---majority vote and reward-model selection plateau as samples grow~\citep{brown2024monkeys}. Closing the loop with the model's \emph{own} judgment, as Self-Refine~\citep{madaan2023selfrefine} and Reflexion~\citep{shinn2023reflexion} do, stalls exactly where self-assessment fails. Our search (\S\ref{sec:verifier}) departs on two axes. First, the feedback is \emph{executed} verification rather than opinion: best-of-$N$ keeps the candidate an executable verifier scores highest, and refinement is a strict verifier-scored hill-climb with backtracking. Second, the operators are wired into the certificate-gated controllers---best-of-$N$ is \alg{4}'s accept-gate pointed at patches, and process rewards feed \alg{3}'s credit assignment. Where model-generated tests have been used for selection and self-debugging~\citep{chen2023codet,chen2024selfdebug}, our verifier (\S\ref{sec:firewall}) keeps a strict firewall between the self-authored oracle that \emph{steers} the search and the held-out grader that only \emph{measures} it, admitting an oracle only if it fails on the unpatched base.

\section{Architecture: Four Layers Around a Frozen Model}
\label{sec:architecture}

The deployed agent at round $t$ is the composition
\begin{equation}
\pol_t \;=\; L_0 \circ L_1^{(t)} \circ L_2^{(t)},
\end{equation}
where the four layers, and the channels through which the five controllers act on them, are summarized in Figure~\ref{fig:arch}:
\begin{description}[leftmargin=1.5em,itemsep=1pt,topsep=2pt]
\item[$L_0$ --- Base model.] A frozen pre-trained LLM, accessed as a function $\mathrm{LLM}(\text{prompt})\to\text{text}$, optionally returning token log-probabilities. Never updated.
\item[$L_1$ --- Adapter.] The trainable policy parameter $\theta$. With open weights this would be a low-rank adapter~\citep{hu2022lora}, prefix~\citep{li2021prefix}, or soft prompt~\citep{lester2021power}; the provider-agnostic realization is a \emph{steering adapter}: a stochastic policy over a finite set of $k$ steering directives, parameterized by logits $\theta\in\R^k$ with $p=\mathrm{softmax}(\theta)$ (\S\ref{sec:substrate}).
\item[$L_2$ --- Harness.] Mutable orchestration: system prompt, tool definitions, step and exploration budgets, memory, the grown abstraction library, and a \emph{repair pipeline} of adopted self-repair primitives (\S\ref{sec:selfrepair}). Edited by \alg{4}; grown by \alg{5}; self-repaired online. Every structural edit changes the harness's identity, so the policy version is well defined at all times.
\item[$L_3$ --- Loop controller.] One of the five algorithms (or their composite). Holds certificates, gates, budgets, and archives; never part of the agent's forward pass.
\end{description}

The performative distribution $\dist(\pol_t)$~\citep{perdomo2020performative} is the distribution of (prompt, tool-call, environment-response) tuples the agent generates in deployment. Each round the controller emits a \emph{certificate}---a structured audit record carrying the round's decision (\textsc{accept}/\textsc{hold}/\textsc{reject}/\NSF), the error budget spent, and algorithm-specific metrics. Certificates form the unified ledger by which a run can be audited after the fact.

\subsection{The policy substrate}
\label{sec:substrate}

The pseudo-code in this paper is written against the following concrete objects.

\paragraph{Steering adapter ($L_1$).} The only trainable parameters are logits $\theta\in\R^k$ over $k$ steering directives---short natural-language strategy instructions appended to the system prompt. Each round the agent samples one directive, $i\sim\mathrm{softmax}(\theta)$, so the distribution over directives is an explicit softmax whose log-probability is known exactly on any backend. \emph{We never differentiate through the frozen model.} The only gradient any controller uses is the closed-form score-function (REINFORCE) gradient of this softmax,
\begin{equation}
\label{eq:score-grad}
\nabla_\theta\log p_\theta(i)=\mathbf{1}_i-\mathrm{softmax}(\theta),
\end{equation}
which needs only the sampled index $i$ and a softmax---no backpropagation, no model internals---and is therefore exact even on a text-only API. This is the gradient \alg{1} and \alg{3} consume. For the forgetting gate the same $\theta$ is read as a diagonal-Gaussian PAC-Bayes posterior $Q_\theta=\mathcal{N}(\theta,\mathrm{diag}\,e^{v})$ with fixed log-variance $v$, so $\KL(Q_\theta\|Q_0)$ is just a scaled squared distance to the prior mean; the same vector supplies the $\ell_2$ distance and ball projection the trust region needs. No model Jacobian is ever formed: the ``Jacobian memory'' of orthogonal-gradient continual learning is approximated by a finite FIFO buffer of past closed-form gradients~\eqref{eq:score-grad} (\alg{1}), and true open-weights Jacobian buffers, though supported, are used in no reported run. Every committed policy is thus identifiable and auditable after the fact.

\paragraph{Policy identity and distance.} A policy's \emph{version} is determined by $(L_1,L_2)$; $L_0$ is fixed. Policy distance is $\lVert\theta-\theta'\rVert_2$ plus a unit structural step if the harnesses differ---harness edits are discrete, and their performative effect is estimated from deployment rewards rather than from a norm.

\paragraph{Deployment.} $\textsc{Deploy}(\pol,\text{tasks})$ runs all tasks concurrently and returns one rollout per task: the directive index used, the action, the tool trajectory, and a bounded reward $R\in[0,1]$ from the environment. Two paths exist behind one interface: a single-call path (one LLM generation per task), and an \emph{actor} path in which a multi-step tool-using agent (\S\ref{sec:composite}) produces the final action. An attempt index varies the sampling seed and the working directory, so independent attempts of the same (policy, task) pair are diverse and isolated---the substrate for best-of-$N$ and refinement search (\S\ref{sec:verifier}). Rollouts are appended to a persistent replay buffer, which \alg{3} and \alg{4} consume off-policy.

\paragraph{Access levels.} What a provider exposes determines what is implementable. \emph{API-text} (text only) already suffices for every controller, because the trainable policy is the $L_1$ softmax and its gradient~\eqref{eq:score-grad} needs only sampled directive indices: \alg{1}, \alg{2}, and \alg{3} run even on local text-only servers, and \alg{4} and \alg{5} edit only $L_2$. \emph{API-logprob} adds sequence-level importance ratios; \emph{open weights} would additionally allow gradients through the model itself and true Jacobian memory---neither of which is used here.

\section{The Five Loop Controllers}
\label{sec:algorithms}

Throughout, $\pol_t$ is the deployed configuration at round $t$; $\dist(\pol)$ the induced data distribution; $\varepsilon$ the performative sensitivity ($\Wone(\dist(\pol),\dist(\pol'))\le\varepsilon\lVert\pol-\pol'\rVert$); $L$ the loss's smoothness-to-curvature ratio (the condition number $\beta/\gamma$ of Thm.~3.5 of \citet{perdomo2020performative}, so the contraction regime is $\varepsilon L<1$); $\delta_0$ a global error budget; and \NSF{} (``no solution found'') the safe abstention output~\citep{thomas2015high,thomas2019preventing}. All rewards are bounded in $[0,1]$; the per-rollout loss is $\ell=1-R$.

\paragraph{Two tiers, ten algorithms.} The five controllers in this section (\alg{1}--\alg{5}) are the conceptual core. The reference implementation pairs them with five verifier-in-the-loop / actor-side mechanisms (\alg{6}--\alg{10}, \S\ref{sec:verifier}--\S\ref{sec:selfrepair}) that supply the signal and variation the controllers act on: gates can only \emph{select} among existing behaviors, so the engine that generates and verifies those behaviors carries its own numbering. Table~\ref{tab:algmap} is the full inventory. Pseudocode for all algorithms is collected in Appendix~\ref{app:pseudocode}.

\begin{table}[t]
\centering
\caption{The implementation's ten numbered algorithms. \alg{1}--\alg{5} are the scheduled $L_3$ controllers of this section; \alg{6}--\alg{10} are the verifier-in-the-loop and actor-side mechanisms of \S\ref{sec:verifier}--\S\ref{sec:selfrepair}. The five controllers remain the contribution; \alg{6}--\alg{10} are the engine that feeds them. \alg{4} is omitted from the live SWE stack for wall-clock cost (\S\ref{sec:composite}) and best-of-$2$ (\alg{6}) is removed from it (\S\ref{sec:ablation}); both remain in the catalog.}
\label{tab:algmap}
\small
\begin{tabular}{@{}llll@{}}
\toprule
Flag & Mechanism & Where & Acts on \\
\midrule
\code{alg1}  & PPB-CL: forgetting-gated continual learning      & \S\ref{sec:alg1}, Alg.~\ref{alg:ppbcl}      & $L_1$ \\
\code{alg2}  & PNMP-A: anchored preference learning             & \S\ref{sec:alg2}, Alg.~\ref{alg:pnmpa}      & reward $q$ \\
\code{alg3}  & PA-COCOA: counterfactual credit assignment       & \S\ref{sec:alg3}, Alg.~\ref{alg:pacocoa}    & $L_1$ \\
\code{alg4}  & SGM-CS: confidence-gated harness edits            & \S\ref{sec:alg4}, Alg.~\ref{alg:sgmcs}      & $L_2$ \\
\code{alg5}  & SDC-QD: MDL+QD library growth                     & \S\ref{sec:alg5}, Alg.~\ref{alg:sdcqd}      & $L_2$ \\
\midrule
\code{alg6}  & verifier-gated best-of-$N$ $+$ refinement         & \S\ref{sec:verifier}, Alg.~\ref{alg:verifier} & search \\
\code{alg7}  & verified micro-step search (fast loop)            & \S\ref{sec:micro}, Alg.~\ref{alg:micro}     & search \\
\code{alg8}  & self-authored reproduction oracles               & \S\ref{sec:firewall}, Alg.~\ref{alg:firewall} & verifier \\
\code{alg9}  & search-layer controllers (\alg{1}/\alg{3}/\alg{4}/\alg{5} re-aimed) & \S\ref{sec:twoloop}, Alg.~\ref{alg:continual} & search \\
\code{alg10} & verified self-repair of the harness               & \S\ref{sec:selfrepair}, Alg.~\ref{alg:repair} & $L_2$ \\
\bottomrule
\end{tabular}
\end{table}

\subsection{\alg{1}: Performative PAC-Bayes Continual Learning (PPB-CL)}
\label{sec:alg1}

\textbf{Problem.} The agent must acquire new skills without catastrophically forgetting old ones, on a stream it generates itself. The two natural seed results do not cover this setting: OGD's forgetting guarantee assumes an exogenously indexed task sequence in the infinite-width regime~\citep{farajtabar2020orthogonal,bennani2020generalisation}, and time-uniform PAC-Bayes forgetting certificates~\citep{chugg2023unified,friedman2025forgetting} require a data-independent prior---which fails the moment the agent reuses its own evolving policy as a prior. The stream, the loss, and the prior are all endogenous here.

\textbf{Solution (Algorithm~\ref{alg:ppbcl}).} \emph{Intuition.} Damp each update by how much it would degrade a small anchor set of past tasks, and cap its size by the remaining forgetting budget. Concretely, keep the PAC-Bayes object small and the prior frozen---the posterior is a mean-parameterized diagonal Gaussian $Q_\theta$ over the steering adapter, with a frozen data-independent prior $Q_0$. Each round, a performative batch is deployed and a REINFORCE gradient of the expected loss is formed from the sampled directive indices, with the baseline pooled over the recent replay window (a within-round baseline is identically zero at batch size one). The gradient is projected orthogonally off a FIFO buffer of the last $m$ committed update directions (the OGD guard), and a candidate adapter is proposed by a single descent step. The candidate must pass three gates in sequence. (i) \emph{Forgetting gate}: the candidate is deployed on a small \emph{anchor set} of past tasks, each stored with its best historical reward $R^*_j$ (with $M{=}1$ for rewards in $[0,1]$; until anchors accumulate the gate passes vacuously); the empirical backward transfer $\widehat{\mathrm{bt}}=\frac{1}{|\mathcal{A}|}\sum_j[(1-R_j)-(1-R^*_j)]$ enters the Donsker--Varadhan bound
\begin{equation}
\label{eq:dv}
B_{\mathrm{fgt}}(\theta) \;=\; \widehat{\mathrm{bt}}(\theta) + \frac{\KL(Q_\theta\|Q_0)+\log(2\sqrt{t}/\delta)}{\lambda} + \frac{\lambda M^2}{8|\mathcal{A}|},
\end{equation}
an instance of the anytime PAC-Bayes recipe of \citet[Thm.~3.1]{chugg2023unified}---which holds for precisely the adapted posterior sequences a self-evolving agent produces---via Ville's inequality; the temperature is set per evaluation to its optimizer $\lambda^*=\sqrt{8|\mathcal{A}|A_t}$ with $A_t=\KL(Q_\theta\|Q_0)+\log(2\sqrt t/\delta)$, collapsing the last two terms of \eqref{eq:dv} to $\sqrt{A_t/(2|\mathcal{A}|)}$ (a fixed $\lambda$ left the bound an order of magnitude above any usable threshold). If $B_{\mathrm{fgt}}>\tau_{\mathrm{forget}}$, the step is damped geodesically toward the incumbent by factor $\rho$ and re-tested, up to five times. (ii) \emph{Performative trust region}: the surviving slack $\tau_{\mathrm{forget}}-B_{\mathrm{fgt}}$, divided by $\varepsilon$, caps the adapter step in $\ell_2$; a violating candidate is projected onto the ball and the bound re-evaluated. This is what keeps the induced shift inside the contraction regime of \citet{perdomo2020performative}. (iii) \emph{Commit gate}: the candidate commits only if $B_{\mathrm{fgt}}\le\tau_{\mathrm{forget}}$ and $\varepsilon L<1$; on commit, the projected gradient joins the buffer and the anchors absorb the round's tasks with their best rewards. The certificate reports the empirical risk, $\KL$, the time-uniform penalty $\sqrt{(\KL+\log(2\sqrt t/\delta))/2n_t}$ and risk bound, the forgetting bound, trust radius, and damping count.

The pieces are published; their composition is an empirical construct we do not prove correct under the endogenous loop. Each ingredient is sound on its own---the time-uniform PAC-Bayes penalty of \citet[Thm.~3.1]{chugg2023unified}, the Donsker--Varadhan backward-transfer forgetting bound of \citet[Thm.~3.1]{friedman2025forgetting}, and the performative contraction regime $\varepsilon<\gamma/\beta$ of \citet[Thm.~3.5]{perdomo2020performative}---but whether they combine into an anytime-valid forgetting guarantee that survives the loop is a conjecture, not a theorem we establish. The capped direction buffer in particular forfeits the exact no-forgetting property of \citet[Thm.~2]{bennani2020generalisation}, which assumes unbounded Jacobian memory; and the anytime floor of \eqref{eq:dv} grows with $t$, so $\tau_{\mathrm{forget}}$ must sit above it or the gate freezes the learner.

\subsection{\alg{2}: Performative Nash-MP with Real-Data Anchoring (PNMP-A)}
\label{sec:alg2}

\textbf{Problem.} Recursive self-training and self-rewarding loops drift to degenerate fixed points: the policy exploits the reward model's over-predictions, and the reward model is retrained on the policy's own outputs~\citep{gao2023scaling,shumailov2024collapse}. The seeds each cover half the problem: constant-fraction real-data anchoring prevents data collapse~\citep{fu2025collapse} but says nothing about reward-model drift, while last-iterate Nash convergence~\citep{tiapkin2025nashmp,wang2025magnetic} assumes a \emph{fixed} preference game. The loop must be protected against both simultaneously.

\textbf{Solution (Algorithm~\ref{alg:pnmpa}).} \emph{Intuition.} Mix a fixed reference anchor into every reward-model update and gate any update that drifts from it, while letting the policy chase the reward model on a slower timescale than the model settles. Two ideas do the work: separate the two risks onto two timescales~\citep{borkar2008stochastic}, and gate the slow one. The preference model is a Bradley--Terry score vector $q\in\R^k$ over directives, $P(i\succ j)=\sigma(q_i-q_j)$, with a frozen anchor $q_{\mathrm{real}}$ (absent a held-out real-preference set the anchor defaults to the neutral zero vector, so in the SWE runs it prevents drift from neutrality rather than from human preferences). Synthetic preference pairs come from directives pooled over the recent replay window, the higher-mean directive winning. The \emph{slow} timescale blends the synthetic Bradley--Terry gradient with a pull toward the anchor,
\begin{equation}
\label{eq:pnmp-slow}
q_{\mathrm{cand}} = q + a^{\mathrm{slow}}_t\big[\alpha\,(q_{\mathrm{real}}-q) + (1-\alpha)\,\nabla_q\widehat{\mathcal L}_{\mathrm{BT}}(q)\big],\qquad a^{\mathrm{slow}}_t=\tfrac12(t{+}1)^{-1},
\end{equation}
and is adopted only if it clears a drift gate: the per-pair deviations from the anchor, $X_{ij}=|\sigma(q^{\mathrm{cand}}_i-q^{\mathrm{cand}}_j)-\sigma(q^{\mathrm{real}}_i-q^{\mathrm{real}}_j)|$, feed a Hoeffding e-process testing $H_0\!:\E[X]\le\tau$, and rejection at the round's CTHS level $\delta_k$ keeps the last safe model. The \emph{fast} timescale moves the policy by a regularized Nash mirror-prox (extragradient) step on the logits,
\begin{equation}
\label{eq:pnmp-fast}
\theta_t = \theta_{t-1} + a^{\mathrm{fast}}_t\,(\theta_{\mathrm{MP}}-\theta_{t-1}),\qquad a^{\mathrm{fast}}_t=\tfrac12(t{+}1)^{-0.7},
\end{equation}
whose game gradient is the self-play advantage $\mathrm{adv}_i(p)=\sum_j p_j\,\sigma(q_i-q_j)$ and whose prox operator folds in the KL to the previous iterate and $\beta$ times the KL to the reference policy (the two extragradient half-steps are spelled out in Algorithm~\ref{alg:pnmpa}). Because $a^{\mathrm{slow}}_t/a^{\mathrm{fast}}_t\to0$, the model settles faster than the policy chases it---the required timescale separation. A magnet $z$ is refreshed every $K$ rounds and $\KL(\pol_t\|z)$ reported on the certificate; once the budget $\delta_0$ is spent the preference model freezes and only policy updates continue.

Here too the components are published and the combination is an unproven, empirical construct. We borrow the constant-$\alpha$ bound on cumulative distribution shift in self-consuming training of \citet[Thm.~1]{fu2025collapse}, the geometric last-iterate self-play contraction (up to a residual floor) of \citet[Prop.~1]{tiapkin2025nashmp}, and the anytime-valid familywise control of e-processes under harmonic spending~\citep{ramdas2023game,wu2025sgm}. Whether their two-timescale coupling converges to a performative equilibrium without collapse is a conjecture, not a result we prove; we claim no rate, and note that performative equilibria need not be unique and that the shift bound degrades as $\alpha\to0$.

\subsection{\alg{3}: Performative-Aware COCOA (PA-COCOA)}
\label{sec:alg3}

\textbf{Problem.} Long-horizon, sparse outcomes must be attributed across a multi-component agent stack while the transition and reward kernels move with the deployed policy~\citep{mandal2023performative}. COCOA's unbiasedness~\citep{meulemans2023cocoa} holds under a fully-predictive outcome encoding---$p^{\pol}(R_k{=}r\mid S_0,A_0,U_k{=}u)=p^{\pol}(R{=}r\mid U{=}u)$ for all $k$ (their Def.~2)---with ground-truth contribution coefficients (their Thm.~1), \emph{in a fixed MDP}; under endogenous drift the policy-induced shift is a hidden confounder, and the high-variance importance ratios of REINFORCE compound over the horizon.

\textbf{Solution (Algorithm~\ref{alg:pacocoa}).} \emph{Intuition.} Replace ``reward $\times$ log-prob'' with a counterfactual per-action contribution, condition on the deploying policy so policy drift stops confounding, and feed the result to a parameter-free learner that restarts on detected drift. Two moves realize this. First, \emph{augment} the outcome encoding with the deploying policy, $\tilde{U}=(U,\pol_t)$, so the shift enters the conditioning set and stops confounding; concretely, every replay rollout carries its policy version and directive index, and the contribution model is estimated per directive over a recency-weighted replay window, with the policy-induced drift absorbed by the recency weighting rather than by explicit conditioning on the version (a deliberate simplification of the full $(U,\pol_t)$ encoder). The contribution model is thus a recency-weighted per-directive mean of episode rewards over the recent window (half-life $h$; weight $2^{-\mathrm{age}/h}$; default $\tfrac12$ for unseen directives)---where the episode reward is the \emph{process reward} $\tilde R$ of \S\ref{sec:verifier} when available, i.e.\ the best verifier state the episode reached rather than only its terminal submit. The augmented COCOA gradient then sums counterfactual contributions over \emph{all} actions, eliminating the importance ratio; for the softmax steering adapter it has the closed form
\begin{equation}
\label{eq:cocoa-grad}
g \;=\; \sum_{a'}\widehat{w}(a')\,\nabla_\theta\,p_\theta(a') \;=\; \big(\widehat{w}\odot p\big) - p\,\langle \widehat{w},p\rangle,\qquad p=\mathrm{softmax}(\theta),
\end{equation}
a simplified instance of the COCOA estimator in which credit attaches to the episode outcome rather than to per-step rewards. A mean-reverting exploration floor adds $\varphi\,(\mathbf{1}/k-p)$ to the gradient, keeping every directive sampled so its credit estimate stays fresh instead of collapsing onto one arm of an inflated signal. Second, hand the update to a learner that needs no learning rate and tracks drift: per-coordinate Krichevsky--Trofimov coin betting~\citep{orabona2016coin} (the iterate is $x=x_0+\beta_t\mathrm{W}_t$ with clipped betting fraction $\beta_t$ and per-coordinate wealth $\mathrm{W}_t$), wrapped with drift-triggered restarts. A seeded wild-bootstrap trend test (\S\ref{sec:stats}) on the mean-reward stream signals non-stationarity and the bettor re-anchors at the current iterate---a restart-based surrogate for the strongly-adaptive guarantees that yield path-length dynamic regret $\widetilde{O}(\sqrt{T V_T})$ on every interval~\citep{cutkosky2020parameter}, and the optimal $\widetilde{O}(T^{1/3}V_T^{2/3})$ rate for strongly convex losses~\citep{baby2022optimal}.

As before, every component is published and the composition is empirical, with no correctness proof in the performative setting. The unbiasedness of the counterfactual-contribution estimator under a fully-predictive encoding is \citet[Thm.~1]{meulemans2023cocoa}, and Proposition~5 there already shows the hindsight model is invariant to conditioning on the policy logits within a deployment; the parameter-free dynamic-regret rates are \citet{cutkosky2020parameter} (path-length $\widetilde O(\sqrt{TV_T})$ on every interval, with $V_T$ the comparator path length) and \citet{baby2022optimal} ($\widetilde O(T^{1/3}V_T^{2/3})$ for strongly convex losses). Whether unbiasedness and these rates carry to the performative MDP~\citep{mandal2023performative} across rounds is a conjecture we do not prove: a fully-predictive $\tilde U$ must be learnable, the recency-weighted contribution estimate reintroduces a bias their theorem does not cover, and the regret statement is vacuous once path variation $V_T$ grows linearly.

\subsection{\alg{4}: SGM Gated by Anytime-Valid Confidence Sequences (SGM-CS)}
\label{sec:alg4}

\textbf{Problem.} A self-modification must be admitted only if it statistically improves the agent, with risk bounded across an \emph{unbounded} sequence of self-proposed edits, where the proposal process is endogenous (it may exploit residual statistical slack) and a deployed edit changes the very distribution in which it is judged.

\textbf{Solution (Algorithm~\ref{alg:sgmcs}).} \emph{Intuition.} Admit a self-edit only through a sequential, peek-safe paired test that spends a shrinking per-edit slice of a global error budget---so familywise error stays bounded over an unbounded edit stream---and correct that test for the distribution shift the edit itself induces; abstain when in doubt. \alg{4} runs on API-text access, edits only $L_2$, and is the governance layer to deploy first. The proposer is the frozen LLM acting on its own harness, cycling the three knobs of the explore/edit economy (\S\ref{sec:verifier}): the forced-edit threshold, the step budget, and the strategy guidance. Baseline $\pol_{t-1}$ and candidate $\pol_{\mathrm{cand}}$ are deployed \emph{concurrently on the same task batch} under common random numbers (aligned per-task seeds, isolated working copies), and the decision is a lower bound on the paired \emph{process}-reward (Eq.~\ref{eq:process}) gain, corrected for performativity:
\begin{equation}
\label{eq:sgmcs-lcb}
\mathrm{LCB} = \mathrm{CS}^{\mathrm{lower}}_{\delta_k}\!\big(R_{\mathrm{cand}}-R_{\mathrm{base}}\big) - \varepsilon\cdot\Wone(R_{\mathrm{cand}},R_{\mathrm{base}}),\qquad \text{admit iff } \mathrm{LCB}\ge-\epsilon_{\mathrm{tol}}.
\end{equation}
The paired-difference confidence sequence is taken at the round's CTHS level $\delta_k=\delta_0/(Z\,k\log^2(k{+}1))$ (\S\ref{sec:stats}); the $\Wone$ term, computed exactly in 1-D from the samples, inflates the bound by the shift the edit induces; and per-version normal-mixture sequences~\citep{howard2021time} accumulate value evidence across rounds. Three details keep the test honest and cheap. (i)~Two free guards spend nothing: an exhausted budget or a no-op proposal simply holds. (ii)~A \emph{pre-gate pilot} on the round's best-evidenced (known-passable) task rejects (\NSF) a candidate that shows no promise there, while an uninformative zero-reward pilot falls through to the full evaluation; the remaining tasks are then ranked so the comparison falls where evidence is weakest. (iii)~A seeded wild-bootstrap trend test~\citep{chandak2020safe}, fed only by rounds that actually re-evaluated the baseline, widens the bound by one radius when the baseline-value stream is non-stationary. Otherwise the round returns \NSF{} and the baseline is kept.

The statistical components are published; their composition under an endogenous proposer is an empirical construct we do not prove safe. We reuse anytime-valid e-processes with familywise control under Ville's inequality~\citep{ramdas2023game}, the confirm-triggered harmonic spending of the Statistical G\"odel Machine~\citep{wu2025sgm}, the time-uniform confidence sequence of \citet{howard2021time}, and the abstention semantics of safe policy improvement~\citep{thomas2015high,thomas2019preventing}. Whether familywise safety survives the endogenous proposal process and the edit-induced distribution shift is open: the performative correction rests on a knowable bound for $\varepsilon$, and the wild-bootstrap widening~\citep{chandak2020safe} was derived for \emph{exogenous} non-stationarity, so under the endogenous loop it is a conservative heuristic. By construction the protocol favors \emph{safety over progress}---it may rationally abstain (\NSF{}) indefinitely. So that this conservatism never discards a good \emph{solution} (as opposed to a good harness \emph{edit}), the SWE instantiation keeps a \emph{shadow-best}: the highest-reward candidate patch seen across rounds is retained even when the harness-edit gate abstains, decoupling ``which harness to deploy'' (gated) from ``which patch to submit'' (best-so-far); and a rejected harness that nonetheless produced a strong patch is \emph{requeued} for a later round, with eligibility keyed on an absolute reward threshold and a bounded per-harness retry count, and replayed only in a round whose active repositories include its own.

\subsection{\alg{5}: Stitch-in-DreamCoder with Quality-Diversity Acceptance (SDC-QD)}
\label{sec:alg5}

\textbf{Problem.} The agent must grow its own hypothesis space---invent reusable abstractions---while the corpus those abstractions are mined from is generated under the very library being learned. DreamCoder's endogenous loop has no convergence or no-collapse theorem~\citep{ellis2021dreamcoder}; Stitch's MDL-optimal abstraction step is one-shot~\citep{bowers2023stitch}. A single converging library is the wrong target when corpus generation is endogenous.

\textbf{Solution (Algorithm~\ref{alg:sdcqd}).} \emph{Intuition.} Keep a quality-diversity archive of behaviorally distinct libraries rather than one ``best'' library, and add an abstraction only when it both reduces description length and occupies a new niche; each addition is the exact most-compressive one on the current corpus. Concretely, replace the single library with an \emph{archive over a frontier}, and make each growth step exact. Programs are S-expressions with description length $|\rho|$ equal to the node count, and the objective is the node-count MDL surrogate $J(L;C)=|L|+\sum_{\rho\in C}|\rho|$ of DreamCoder's description-length posterior over libraries~\citep{ellis2021dreamcoder}. Each round, a \emph{wake} phase asks the frozen LLM to synthesize one program per task under the current library (shown in the prompt; output is parsed defensively, since LLM output may be prose); programs whose environment score reaches the solve threshold join the corpus. The \emph{sleep} phase runs Stitch-style compression: candidate patterns are all corpus subtrees plus all pairwise antiunifications of structurally compatible subtrees (differing positions become numbered holes), and the net utility of a pattern with body size $b$, arity $a$, and $m$ non-overlapping matches is $u=m\,(b-1-a)-b$ (each match is rewritten as an application of size $1+a$; the definition is paid once). Candidates are scanned in descending order of utility with an early dominance break, so the maximizer over the candidate set is exact (the soundness argument behind Stitch's dominance pruning, Lemma~1 of \citet{bowers2023stitch}). The best abstraction $A^*$ is accepted only if (i) $\Delta J=u(A^*)\ge u_{\min}$, the minimum-utility bar (so a marginally compressive pattern does not enter the library), and (ii) the resulting library is \emph{novel or improving} in a MAP-Elites archive keyed by the behavior descriptor $\phi=(\text{library size},\text{mean solved-program size})$ with fitness the post-compression MDL---so the run illuminates a (compression, coverage) frontier rather than converging to one point~\citep{mouret2015illuminating,cully2018quality}. Accepted abstractions are surfaced as $L_2$ harness entries the agent can call. A Hoeffding-type per-task description-length certificate, in the spirit of the PAC-Bayes lifelong-learning bound of \citet{pentina2015pacbayesian}, bounds expected per-task description length on held-out tasks:
\begin{equation}
\label{eq:catoni}
\E_{\text{new task}}[\mathrm{DL}] \;\le\; \frac{J(L_t;C_t)}{|C_t|} + \frac{|L_t|}{|C_t|} + \sqrt{\frac{\log(1/\delta)}{2|C_t|}}.
\end{equation}
On the SWE agent (\S\ref{sec:composite}), the wake phase is replaced by the agent's own behavior: successful trajectories, encoded as $(\text{tool},\text{target})$ operation sequences, become the corpus, and the mined macros become callable harness entries.

The per-round compression step is exact and published; the loop around it is an empirical construct without a convergence proof. The single step is exact in the published sense: Lemma~1 of \citet{bowers2023stitch} shows that dominance pruning never discards the optimal abstraction, so the accepted macro is the most-compressive one on the current corpus (over the implemented candidate set), and MDL is monotone non-increasing under a fixed corpus. The held-out description-length certificate (Eq.~\ref{eq:catoni}) is Hoeffding/Occam-type---it carries no $\KL$ term and so is not itself a PAC-Bayes bound, but shares the Hoeffding-lemma core of the lifelong PAC-Bayes bound of \citet{pentina2015pacbayesian}---and assumes an exchangeable task distribution that endogenous generation may violate. Convergence of the quality-diversity archive to a meaningful frontier is \emph{not} a published result---no QD framework, including \citet{cully2018quality}, proves it---so we make no such claim; it is at most a conjecture, and the global limit need not be meaningful when corpus generation is endogenous. Every abstraction is a composition within a fixed typed $\lambda$-calculus; expanding the type system is beyond this algorithm.

\section{The Verifier as an In-Loop Control Signal}
\label{sec:verifier}

\textbf{Problem.} The five controllers gate, select, and reshape: they act on the distribution of behaviors the frozen base model already produces. On hard agentic tasks that distribution offers a passive learner little. (i) The natural reward is \emph{terminal and binary}---an instance is resolved or it is not---so a base that never fully resolves yields an identically zero signal and the loop has no gradient. (ii) Some failures are \emph{systematic, not stochastic}: on certain instances a base re-runs the \emph{same} attempt under temperature, returning identical partial patches and never the occasional good one, so selection over samples has no variance to exploit (an instance-level phenomenon, not a property of a whole model). (iii) Yet the same base \emph{is} reliable at the \emph{micro-step}---it can localize and emit an applying one-line edit---even when it gets the whole-patch \emph{content} wrong; reliability lives at a finer granularity than the unit the loop scores. (iv) The interventions that helped were \emph{control-flow}, not capability: forcing an edit after an exploration budget changes what the model does next, whereas coaxing latent quality does not. (v) The verifier---the one component that knows ground truth---was consulted once, after the episode, when nothing could be done about what it reported; this is exactly the regime in which selection, not generation, is the bottleneck~\citep{brown2024monkeys}. The problem is to convert the verifier from a passive end-of-episode score into a control signal that acts \emph{inside} the episode, \emph{across} attempts, over the \emph{action space}, and inside \emph{credit assignment}, and to relocate both the search and the reward to the micro-step granularity where the base is reliable---without ever leaking gold labels.

\textbf{Solution.} A graded verifier and a family of search operators, all derived from test execution alone and all composable with every controller through the actor interface (the attempt-indexed deployment substrate of \S\ref{sec:substrate}), building up to a verified micro-step search (\S\ref{sec:micro}).

\paragraph{Graded verifier.} A shaped milestone reward first decomposes the path to a resolve into rungs derived only from test execution: $0$ for no parseable diff, $0.2$ for a diff that does not apply, and $0.4+0.6\cdot\mathrm{frac}_{\mathrm{f2p}}$ once tests run (halved and capped at $0.55$ when pass-to-pass regressions appear), capped at $1$. Because a whole batch of edits can sit at the same rung (every one applies but none passes a test), this is still too coarse to climb; the \emph{graded} verifier adds two finer signals on top of the milestone score $s_{\mathrm{shaped}}$:
\begin{equation}
\label{eq:graded}
V \;=\; \begin{cases}1 & \text{resolved}\\[2pt]
\min\!\big(0.99,\ s_{\mathrm{shaped}} + \mathbb{1}[s_{\mathrm{shaped}}\ge0.4]\,(w_p\,\mathbb{1}[\text{progressed}] + w_j\,j)\big) & \text{otherwise,}\end{cases}
\end{equation}
where the bonuses apply only once a patch has cleared the apply-and-run rung ($s_{\mathrm{shaped}}\ge0.4$), \emph{progressed} is true when the test's failure \emph{signature} moves (the last \code{pytest} error line, with volatile numbers and addresses masked, differs from the seed's---so the edit got further before failing), $j\in[0,1]$ is an auxiliary LLM judge, and the bonuses $w_p,w_j$ are kept small ($0.08$ each) so an actual sub-test pass always outranks the heuristics and a non-resolved patch never reaches $1$. Error-progression is the key addition: it turns a plateau of equally-applying patches into a gradient. The fail-to-pass fraction $\mathrm{frac}_{\mathrm{f2p}}$ here is computed from each instance's held-out gold tests, so this graded score is reserved for \emph{terminal measurement} (and offline development), exactly as a test set should be---never to steer the search. The dense signal the search actually climbs is supplied by a self-authored verifier that reads only the issue (\S\ref{sec:firewall}).

\paragraph{Closed-loop test execution.} A \code{run\_tests} tool lets the agent run the real target tests on its \emph{current} edits mid-episode and read back actionable feedback: pass counts and the tail of the failure log. A short, \emph{typed repair instruction} keyed to the failure class (a name or import error, an assertion mismatch, a syntax error, or a no-patch episode each map to a specific corrective hint) is injected into the guidance of the next \emph{refinement attempt} (Algorithm~\ref{alg:verifier}), pairing each recorded failure with a concrete next step. The tool is stateful (its result depends on the current edits), so it is exempt from the agent's duplicate-action guard. This changes the task from one-shot patch guessing to closed-loop debugging---the agent can edit, test, read a failure paired with a concrete next step, and iterate before submitting.

\paragraph{Process reward.} With the verifier callable mid-episode, credit assignment need not wait for the terminal submit. The process reward of an episode is the best verifier state it reached,
\begin{equation}
\label{eq:process}
\tilde R \;=\; \max\!\big(R_{\mathrm{terminal}},\ \max_j s_j\big),
\end{equation}
where $s_j$ is the shaped score at the agent's $j$-th \code{run\_tests} call. A trajectory that reached a good state and then regressed still carries credit; \alg{3}'s contribution model (Algorithm~\ref{alg:pacocoa}) consumes $\tilde R$ in place of the terminal reward.

\paragraph{\alg{6}: Verifier-gated best-of-$N$ and refinement search (Algorithm~\ref{alg:verifier}).} Across attempts, two search operators sit on top of any actor. \textsc{BestOfN} runs $n$ independent, diverse attempts (the attempt index varies the sampling seed and the working directory) and keeps the one the verifier scores highest: under independent attempts, a per-attempt resolve probability $p$ becomes $1-(1-p)^n$---a strict lift whenever $p>0$. Empirically, coverage under repeated sampling grows along an approximate exponentiated power law in $n$, and a ground-truth verifier is exactly the selector that converts coverage into solved instances~\citep{brown2024monkeys}. \textsc{Refine} is a sequential hill-climb: each attempt after the first is seeded with the best patch so far \emph{and the verifier's explanation of why it failed}, the new candidate is scored on the original task, and it replaces the incumbent only if strictly better---otherwise the search backtracks to the best known patch. Refinement stops early on a full resolve. We \emph{removed} best-of-$2$ from the live stack: the run logs show it never produced a second attempt (the attempt count stays $1$) and, run in isolation, degraded the apply-rate, so it is reported here as a designed operator rather than a live component (\S\ref{sec:ablation}). Selecting by the verifier is \alg{4}'s accept-gate pointed at patches; keeping the best over diverse samples is the selection principle of \alg{5}---here applied at test time, within a single task.

\paragraph{Enforced explore$\to$edit budget.} Over the action space, the harness gains a tunable threshold $b_{\mathrm{explore}}$: once the agent has spent $b_{\mathrm{explore}}$ read-only steps (\code{search}/\code{read}/\code{list}) without landing an edit, it is first warned to edit and then, after a short grace window of a few further explores under escalating pressure, the exploration tools are disabled until an edit is made. The grace window is not cosmetic: a hard cap with no grace deadlocked a mid-localization agent into a no-op read loop that produced an empty patch, whereas the graded pressure let the same agent land its edit (in one live case turning an empty patch into a correctly localized fix). This converts the diagnosed over-exploration failure from a property of the model into a property of the harness---and the harness is governed: $b_{\mathrm{explore}}$ is a knob in \alg{4}'s proposer cycle (Algorithm~\ref{alg:sgmcs}), so tightening it must survive the confidence-sequence gate like any other self-edit.

\subsection{\alg{7}: Verified micro-step search (the fast loop)}
\label{sec:micro}

Best-of-$N$ and refinement search over \emph{whole patches}, where a weak base is unreliable and near-deterministic. The fast loop (Algorithm~\ref{alg:micro}) instead searches over \emph{verified micro-steps}: from the current best patch and the verifier's failure feedback, a \emph{reasoning-first} generator (first name the root cause, class, and method; then emit edits) proposes $k$ materially different one-line hypotheses, diversity forced by the prompt against a working memory of past failures rather than by sampling temperature. Each hypothesis is composed onto the current best and scored only if it applies. Three components keep the search efficient: a beam of width $b$ over verified partial patches; the working memory of tried edits and distinct failure signatures, rendered into the prompt to suppress repeats; and a cheap-to-expensive verification cascade (a parse/apply check, then the native fail-to-pass run only on survivors) that confines the minute-long test run to compiling patches. The search halts on a pass and gives up when a full round of fresh hypotheses moves nothing. Hypotheses may be pooled across several base models into one verified beam.

\subsection{\alg{8}: The in-loop verifier --- self-authored reproduction oracles}
\label{sec:firewall}

\textbf{Problem.} The search needs a dense, per-candidate signal to climb, but the held-out gold tests must be reserved for unbiased terminal measurement---a learner that is allowed to consult its own test set will report progress that does not generalize. So the in-loop verifier must be computable from the \emph{issue alone}, using no information from the gold tests, the test patch, or the fail-to-pass/pass-to-pass lists, while still being informative enough to tell a real fix from a near-miss.

\textbf{Solution (Algorithm~\ref{alg:firewall}).} \emph{Intuition.} Reproducing a described bug is easier than fixing it, so the model writes tests from the issue, keeps only those that fail on the unpatched code, and debugs against them; the held-out tests are touched only at the end, to report whether the self-authored verifier was right. Concretely, a verifier model reads only the issue text (and, optionally, repository source) and writes minimal scripts that assert the \emph{correct} expected behavior, so each script should fail on the buggy code and pass once the bug is fixed. The leverage is asymmetric: reproducing a described bug is a higher-success task than fixing it---the issue states the symptom, the model need only encode it---so a model unreliable at fixing can still author a usable verifier.

\emph{Admission.} An oracle is kept only if it \emph{fails on the unpatched base} (a check that already passes on the buggy code captures nothing), with two cheap rejections: timeouts, and failures that are unambiguously the script's own fault, i.e.\ a syntax or indentation error. An import or name error is deliberately \emph{not} auto-rejected---a bug whose very symptom is a broken import must remain admissible---and a symptom judge then demotes any oracle whose base failure does not match the issue's stated symptom. To improve \emph{recall} (whether a usable oracle gets built at all), oracles are pooled across an \emph{ensemble} of verifier models, so a verifier exists whenever \emph{any} model authors one; this only trades precision in a safe direction, since a spurious oracle can make the score \emph{under}-claim (a correct patch fails to flip an over-strict check) but can never mark a wrong patch correct.

\emph{Extracting the signal.} Let $A$ be the admitted oracle suite. A candidate patch $\rho$ is run once (prepared a single time, then every oracle executed against it), and its scalar score is the fraction of admitted oracles it flips from failing to passing, voided if any previously-passing check regresses:
\begin{equation}
\label{eq:vself}
V_{\mathrm{self}}(\rho) \;=\; \frac{\big|\{\,o\in A : o \text{ fails on base},\ o \text{ passes on } \rho\,\}\big|}{|A|}\quad\text{if no green check regresses, else } 0 .
\end{equation}
A patch is \emph{promising} (we never say ``resolved'') when every admitted oracle flips and nothing regresses. Here $V_{\mathrm{self}}$ is the flip fraction (the regression-voiding term is optional), and the held-out grader runs the fail-to-pass directives, so ``the grader agrees'' means the gold target tests flip. The searcher is the multi-step tool-using agent, whose \code{run\_tests} tool returns $V_{\mathrm{self}}$ rather than the grader, so it debugs against a signal it authored; a robust search-and-replace applier (an edit's anchor must match exactly once, with a unique whitespace-normalized fuzzy fallback and a nearest-match hint on a miss) turns the model's edits into a clean diff so even a weak solver emits applying patches without silent wrong-site edits. The held-out grader is invoked only once, on the finalized patch, to measure whether $V_{\mathrm{self}}$ was right. Around the single admission rule the implementation adds three safeguards: rejected oracles are \emph{refined} for a bounded number of rounds (each re-prompt carries the reason for rejection, stopping at the first admission); a quality gate demotes weak oracles whose assertions are negative-only; and oracle code must ground itself in symbols actually extracted from the issue text, with a stoplist rejecting generic tokens.

\paragraph{The no-oracle path.} When no oracle is admitted, the only gold-free evidence is that related pre-existing tests still pass---necessary, not sufficient. This path is hardened in three ways. A fully green but unverified verdict is \emph{capped} at $0.9$ with an explicit ``nothing verified the issue itself is fixed'' note, so best-of-$N$, refinement, verify-react, and the cross-round memory stay armed exactly where verification signal is weakest. Candidates that tie at the cap are split by a gold-free \emph{issue-fidelity judge} (issue text plus the two patches, one-token verdict). A regression veto keyed on the failing phrase prevents an oracle pass from masking a broken pre-existing test, and a collection-error parser closes the remaining false-green channel (a test run that dies during collection no longer reads as green).

\textbf{Why measurement stays honest.} Because $V_{\mathrm{self}}$ is a function only of the issue text and the repository's own behavior, the held-out tests play no role in steering the search; their single terminal call measures the result without having shaped it. The price is that $V_{\mathrm{self}}$ is \emph{fallible}---it can call a patch promising that the grader rejects, or miss a fix its oracles did not cover---which is exactly why a patch is reported as ``promising,'' never ``resolved,'' and why \emph{agreement} between the self-oracle verdict and the independent terminal grader (\S\ref{sec:results}) is the quantity we report.

\subsection{From search to weights: the slow loop and re-aimed controllers}
\label{sec:twoloop}

The fast loop raises quality at test time; a \emph{slow loop} would amortize it into weights by collecting verified micro-step trajectories (label-free positive data---they passed the tests) and distilling them into a low-rank adapter by reject sampling~\citep{zelikman2022star,gulcehre2023reinforced}. \emph{The slow loop is designed but not trained in any reported run} (\S\ref{sec:limitations}). The five controllers are \emph{re-aimed} from the single-pass actor onto this scaffold (Table~\ref{tab:reaim}); four act on the search layer, reuse the primitives of \S\ref{sec:stats}, wrap the beam search of Algorithm~\ref{alg:micro} and the continual stream of Algorithm~\ref{alg:continual}, and are validated \emph{only} by deterministic offline gate simulations, not live:
\begin{itemize}[leftmargin=1.4em,itemsep=1pt,topsep=2pt]
\item \alg{4} becomes a \emph{compute-allocation} controller: an anytime-valid confidence sequence on the per-round marginal graded gain stops a branch once even its optimistic bound clears no further expected gain (a rigorous campaign-level allocator that bounds false-abandonment, plus a cheap per-instance plateau fast-path), reclaiming the expensive verifier budget for productive instances.
\item \alg{5} becomes the \emph{branch diversity generator}: a MAP-Elites archive over a behavior descriptor keeps one elite per behavior cell, so the beam spends its width on behaviorally distinct partial patches rather than $b$ minor variants of one edit---the direct cure for the no-variance failure.
\item \alg{3} becomes \emph{step-level credit}: a recency-weighted mean of \emph{verified} reward per branch class (the edited file) orders which step macro to try first, so the search stops re-discovering the same productive edit shape on every problem.
\item \alg{1} becomes an anchored \emph{forgetting gate}: an update to the evolved search/distilled policy is accepted only if a re-evaluation of frozen earlier-repo tasks does not regress.
\end{itemize}
The fifth, \alg{2}, guards the \emph{slow loop}: it is the mode-collapse defense for the distillation step (an e-value drift gate on trajectory diversity), and is realized only once the slow loop trains---which the present system does not yet do, so \alg{2} at the search layer remains design.

\begin{table}[t]
\centering
\caption{The five controllers re-aimed from the single-pass actor onto the two-loop (search-then-distill) scaffold. Four are built on the search layer and validated by deterministic offline gate simulations; the slow-loop anti-collapse guard awaits the distillation training step.}
\label{tab:reaim}
\small
\begin{tabular}{@{}llp{6.3cm}l@{}}
\toprule
Controller & Acts on & Re-aimed role in the two-loop scaffold & Status \\
\midrule
\alg{4} SGM-CS  & fast loop & anytime-valid \emph{compute-allocation stopping}: halt a branch/instance when the confidence sequence on marginal progress shows no expected gain & built, gate-sim \\
\alg{5} SDC-QD  & fast loop & quality-diversity archive as the \emph{branch diversity generator}; mine step-level macros from verified sub-sequences & built, gate-sim \\
\alg{3} PA-COCOA & fast loop & \emph{step-level} credit over branch / hypothesis classes, off-policy from the verified-trace buffer & built, gate-sim \\
\alg{1} PPB-CL  & slow loop & anchored \emph{forgetting gate}: accept a policy update only if earlier-repo scores hold & built, gate-sim \\
\alg{2} PNMP-A  & slow loop & guard the distillation loop against mode-collapse (e-value drift gate on trajectory diversity) & design (awaits training) \\
\bottomrule
\end{tabular}
\end{table}

\subsection{\alg{10}: Verified self-repair of the harness}
\label{sec:selfrepair}

\textbf{Problem.} A search is only as good as its operators. In the fast loop, a recurring \emph{harness} failure---roughly half of the pooled ensemble's candidate edits failed to apply, because the model copied \code{search} text from the issue's \emph{diff} (lines marked $+$/$-$) rather than from the current file---starves the search of valid candidates. This is exactly the kind of systematic failure a self-evolving agent should fix \emph{itself}. The tempting response is a human hand-patch (``strip the diff markers from \code{search}''); but a hand-patch is an unverified belief about the failure's cause, and acting on it is precisely the trust the rest of the paper forbids.

\textbf{Solution (Algorithm~\ref{alg:repair}).} The harness carries an evolvable \emph{repair pipeline}---a list of adopted repair primitives, an $L_2$ component like \alg{5}'s grown library---applied to every hypothesis before it is composed. Providing a \emph{repertoire} of small, composable primitives (de-marking a diff into verbatim search/replace, stripping code fences, whitespace-fuzzy matching against the actual file) is not hand-patching: the loop \emph{selects and verifies} which to activate. The gate (\textsc{ProposeRepairs}) takes the candidates that just failed to compose and, for each library primitive, \emph{measures} the fraction it makes actually apply against the real composer, adopting only those clearing a threshold, best-first; a repair is credited only when it genuinely changes the hypothesis, so an inert primitive earns nothing. A second, generation-level gate verifies prompt-suffix repairs by \emph{re-generating} and re-composing, adopting a suffix only if it beats the un-amended apply-rate by a margin. Both gates ground adoption in execution, never in a prior. The grown repertoire is the natural target of \alg{5} (which can add primitives) and the per-instance selection is the natural target of \alg{4}'s confidence-gated acceptance.

The mechanism earned its keep by \emph{refusing} a fix: run over the real recorded failures, the diff-de-marking primitive cleared the gate on $0$ of $34$ candidates---the dominant failure was not naive diff prefixes but \code{search} text targeting code the seed patch had already replaced, which no text rewrite can recover---so the verification gate rejected the hand-patch a human would have shipped, and instead adopted a generation-level repair where re-generation raised the measured apply-rate ($33\%\to50\%$ on one anchor) and declined it where re-measurement showed no gain ($83\%\to83\%$ on another). These are single-instance observations, not a controlled evaluation.

Whether each mechanism lifts a given base model---and which failures are stochastic (where best-of-$N$ and refinement have variation to exploit) versus systematic (where the budget enforcement, micro-step relocation, and self-repair bind)---is what the protocol of \S\ref{sec:experiments} measures; \S\ref{sec:results} reports the study that has concluded.

\section{The Anytime-Valid Statistical Core}
\label{sec:stats}

Every controller monitors its own running statistics and may stop, accept, or restart at any round. The shared danger is therefore the same throughout: a fixed-sample test is invalid the moment a controller peeks at its own numbers. The primitives below are the standard tools that stay valid under such continuous peeking; we lead each with what it buys the loop, then give the formula.

\paragraph{Normal-mixture confidence sequences---an error bar valid at every round.} A confidence sequence is an interval that holds for \emph{all} $n$ at once, so a controller can read it after every round without inflating its error rate. For $\sigma$-sub-Gaussian increments with partial sum $S_n$ and intrinsic time $V_n=n\sigma^2$, the mixture supermartingale $M_n=\sqrt{\rho/(\rho+V_n)}\exp\{S_n^2/2(\rho+V_n)\}$ gives, through Ville's inequality, the time-uniform radius
\begin{equation}
\frac{|S_n|}{n}\;\le\;\frac{1}{n}\sqrt{(\rho+V_n)\Big(2\log\tfrac{2}{\alpha}+\log\tfrac{\rho+V_n}{\rho}\Big)}\qquad\text{simultaneously for all }n,
\end{equation}
with $\rho$ tuned to tighten the boundary near a target sample size~\citep{howard2021time}. (The $2\log(2/\alpha)$ is a conservative union of two one-sided boundaries; the exact two-sided boundary of \citet[Eq.~14]{howard2021time} carries $2\log(1/\alpha)$.) This is the value evidence \alg{4} accumulates per policy version.

\paragraph{Hoeffding e-processes---a bankroll that detects drift.} An e-process is a wealth process that grows in expectation only when the null is false, so a controller can bet against ``no drift'' and call drift the moment the wealth crosses $1/\delta$, safely at any stopping time. Concretely, $\prod_i\exp(\lambda_i X_i-\lambda_i^2/8)$ with predictable $\lambda_i$ is an e-process and rejection at $E_t\ge1/\delta$ is anytime-valid~\citep{ramdas2023game}; the bet is a GROW-style plug-in that tilts $\lambda$ toward the observed mean excess using only past data, so it grows fast when drift is real. This is \alg{2}'s reward-model drift gate.

\paragraph{Horizon-free confirm-triggered harmonic spending (CTHS)---one error budget split over unboundedly many edits.} To keep familywise error below $\delta_0$ across an open-ended stream of accepted edits, the per-edit budgets must sum to $\delta_0$. The SGM schedule $\delta_t=\delta_0/(t\,H_B)$, $H_B=\sum_{i\le B}1/i$, achieves this but pre-commits a finite horizon $B$~\citep{wu2025sgm}, which a self-evolving agent lacks. A horizon-free schedule must stay summable on its own: the obvious choice $\delta_0/(2k\log^2(k+1))$ in fact over-spends ($\sum_k\approx1.69\,\delta_0$), silently breaking validity, so the $k$-th confirmation instead spends
\begin{equation}
\delta_k=\frac{\delta_0}{Z\,k\log^2(k+1)},\qquad Z=\sum_{j\ge1}\frac{1}{j\log^2(j+1)}\approx3.39,
\end{equation}
with $Z$ evaluated by an analytic tail correction so that $\sum_k\delta_k=\delta_0$ exactly---familywise validity with no horizon.

\paragraph{Parameter-free coin betting with restarts---an optimizer with no learning rate.} \alg{3} must track a moving optimum without a tuned step size, so it casts each update as a betting game. A per-coordinate KT bettor~\citep{orabona2016coin} keeps wealth $\mathrm{W}_t$ and betting fraction $\beta_t=\frac1t\sum_{i<t}c_i$ (clipped to $\pm\tfrac12$) with normalized reward direction $c_i=-g_i/G_i$ ($G_i$ a running Lipschitz estimate), playing $x_t=x_0+\beta_t\mathrm{W}_t$; this gives comparator-adaptive $O(\lVert u\rVert\sqrt{T\log T})$ regret with no learning rate to set. Drift re-anchors $x_0$ at the current iterate---a restart surrogate for the strongly-adaptive methods that attain path-length dynamic regret on every interval~\citep{cutkosky2020parameter} and the optimal $\widetilde O(T^{1/3}V_T^{2/3})$ rate for strongly convex losses~\citep{baby2022optimal}.

\paragraph{Exact 1-D Wasserstein and sensitivity estimation---how far an edit moved the distribution.} The performative correction (\alg{4}) needs the distance between the reward distributions before and after an edit; in one dimension this $\Wone$ is exact, computed by integrating the gap between empirical CDFs over the merged support. The sensitivity $\varepsilon$ is then the non-negative through-origin slope of $\Wone$ shifts against adapter-norm deltas on probe pairs, and the contraction check $\widehat{\varepsilon}L<1$ is exposed to the controllers.

\paragraph{Wild-bootstrap trend test---is the baseline drifting under us?} Before trusting a logged value stream, the loop tests it for a trend by re-signing the OLS residuals (around a no-trend fit) with Rademacher $\pm1$ multipliers~\citep{chandak2020safe}; the test is robust to heteroskedasticity and heavy tails and is seeded for reproducibility. \alg{3} restarts and \alg{4} widens its bound when it fires.

\paragraph{Weighted importance sampling---reusing old rollouts.} To value a new policy from rollouts logged under older ones, self-normalized weighted IS with clipped log-ratios and an effective-sample-size diagnostic reweights the history, and per-trajectory IS-weighted returns are shaped so a confidence sequence wraps them directly.

\paragraph{MDL compression and QD archives.} Finally, \alg{5} draws on Stitch-style antiunification compression (\S\ref{sec:alg5}) and a MAP-Elites grid archive with coverage, best-elite, and Pareto-frontier queries.

\section{Composite Two-Timescale Control and the Self-Evolving SWE Agent}
\label{sec:composite}

\paragraph{Actor protocol.} The controllers are written against a single-deployment interface; an \emph{actor} protocol lets the same controllers drive a multi-step, tool-using agent: the actor turns (policy, task, attempt) into a final action plus the full tool trajectory and the episode's process reward, recording which $L_1$ strategy directive was sampled (controllers may force a fixed directive---\alg{4} compares harnesses under directive $0$---or let the actor sample on-policy, as \alg{3} requires). Every controller then works unchanged with the multi-step agent, and the search operators of \S\ref{sec:verifier} wrap any actor without modification.

\paragraph{The SWE agent.} The evolvable actor is a multi-step, tool-using agent (ReAct-style~\citep{yao2023react}) over a structured action vocabulary (Appendix~\ref{app:tools}): \code{list}, \code{search}, \code{read}, \code{edit} (unique search-and-replace with a whitespace-fuzzy fallback), \code{run\_tests}, \code{submit}, executed against a working copy at the pre-fix revision with capped observations. Accumulated edits form the candidate patch, so it applies by construction. It reads strategy guidance, the grown macro library, step budget, and exploration threshold from the harness, with the sampled $L_1$ directive composed into the system prompt; each attempt's working copy and best-result archive are keyed by (policy version, directive, attempt), so \alg{4}'s concurrent baseline/candidate arms and best-of-$N$ attempts stay isolated.

\paragraph{Composite controller (Algorithm~\ref{alg:composite}).} A scheduler runs any subset of the five over \emph{one shared evolving policy}: each sub-controller is wrapped as (name, controller, period) and runs when $t\bmod\text{period}=0$; period $0$ disables it---the ablation mechanism. Before a sub-controller steps, it receives the current shared policy; after, its (possibly edited) policy is propagated to the next. Each algorithm edits only its own layer through operations that preserve the other layer, so the edits compose. Sub-certificates are merged into one ledger row: metrics are prefixed per algorithm, error spends are summed, and the round's decision is the highest-precedence sub-decision (\textsc{accept} $\succ$ \NSF{} $\succ$ \textsc{reject} $\succ$ \textsc{hold}). Default periods realize the two-timescale split: $L_1$ controllers (\alg{3} or \alg{1}) every round; $L_2$ controllers (\alg{4}, \alg{5}) every 2 rounds; \alg{2} every 3. \alg{1} and \alg{3} both train $L_1$, so enabling both together is discouraged. An optional \emph{cost-aware} mode makes the scheduler itself a compute allocator: it tracks a per-controller no-gain counter, reset only on \emph{committed} output (an accepted edit, a shadow-accepted patch, a committed macro, or mined anti-macros---raw activity such as solve counts or uncommitted compression gain never trips it), and temporarily skips a slow $L_2$ controller that has shown no committed gain for a few consecutive rounds. A skipped controller is left one round short of the threshold, so a dead controller settles into a probe/skip alternation---a scheduler-level echo of \alg{4}'s per-search compute-allocation stop.

\paragraph{Division of labor.} On the SWE agent the controllers specialize. \alg{4} proposes and gates harness edits (force-edit threshold, step budget, strategy guidance) against the paired process reward (Eq.~\ref{eq:process}) at CTHS levels, with the pre-gate pilot, common-random-number pairing, and shadow requeue of \S\ref{sec:alg4}. \alg{5} mines recurring $(\text{tool},\text{target})$ sub-sequences from high-reward trajectories into macros (and failure signatures into anti-macros), admitting a macro only if it clears a \emph{downstream-utility} gate---positive newer-vs-older reward lift on its context---not merely an MDL gain; the library is capped per (repository, status) context and self-prunes by a trailing-window retirement pass, so the certificate reports compression gain only for committed macros. \alg{3} performs counterfactual credit assignment over rollouts to steer the directive adapter (directives such as \emph{explore broadly}, \emph{localize and edit early}, \emph{smallest diff}), weighting credit per active repository family and excluding \alg{4}'s rejected-candidate arms. \alg{1} protects the adapter with the forgetting gate; \alg{2} anchors any preference model. The cost-aware skip (above) matters here because a scheduled \alg{4} round costs $\sim$90 vs.\ $\sim$7 min for the rest of the composite combined (three concurrent multi-step deploys).

\section{Experimental Setup}
\label{sec:experiments}

\paragraph{Benchmark and grading.} SWE-bench Verified---the 500-instance human-validated subset~\citep{openai2024verified} of SWE-bench~\citep{jimenez2024swebench}---evaluated with the official execution-based harness: for each instance the repository is restored to the state preceding the fix, the candidate patch is applied, and the project's real test suite is run. \emph{Resolved} requires every fail-to-pass test to pass and every pass-to-pass test to keep passing. The working sample is a seeded random draw of $52$ instances ($24$ Django, $27$ Matplotlib, $1$ Flask).

\paragraph{Rewards.} Three bounded rewards are available to the loop. \emph{Proxy}: file-overlap F1 between predicted and gold patches (inner-loop development only, never for claims). \emph{Native}: real test execution, the fraction of fail-to-pass tests passing. \emph{Shaped}: the dense milestone reward of \S\ref{sec:verifier}. The signal live runs actually climb is the self-authored verifier of \S\ref{sec:firewall}; the gold tests are reserved for terminal grading.

\paragraph{Base models and backends.} The study crosses \emph{four} frozen base models, accessed only as $L_0$; none is fine-tuned. Two are open-weight checkpoints served locally with their separate reasoning channel disabled for structured output: \textsc{Gemma} (\code{gemma4:31b-mlx}, a $\sim$31B model at \code{nvfp4} quantization) and \textsc{Qwen} (\code{qwen3.6:27b}, $\sim$27B). Two are frontier reasoning models behind the same interface: \textsc{Gpt-mini} (\code{gpt-5.4-mini}, low reasoning effort) and \textsc{Gpt} (\code{gpt-5.5}, reasoning disabled---the strongest base in the set), each with a $4$k-token completion cap. A fifth model, \textsc{Glm}~5.2, is run with the no-op control and the full suite (Table~\ref{tab:4x2}); its single-pass baseline was not run, so its ``off'' cell is the control. Adapter log-probabilities are exact on every backend because the steered policy is the $L_1$ directive layer.

\paragraph{The $4\times2$: four models $\times$ algorithms off/on.} Each base model is run with the algorithm stack \emph{off} and \emph{on}, on all $52$ instances, under the identical harness and the same official execution-based grader---eight cells, one grading protocol. \emph{No algorithms} is the bare multi-step actor of \S\ref{sec:composite} alone---no controllers, no search, no in-loop test runner---one episode per instance under a $10$-step budget. \emph{Algorithms--A} (the full composite) enables eight of the ten algorithms of Table~\ref{tab:algmap}: \alg{1}/\alg{2}/\alg{3}/\alg{5} scheduled as certificate controllers and the verifier-tier mechanisms \alg{7}--\alg{10} active---verified micro-step search (\alg{7}), the self-authored oracles of \S\ref{sec:firewall} at two samples (\alg{8}) with the unverified-green cap and tie-break judge, search-layer step-credit (\alg{9}), and verified self-repair (\alg{10})---with \alg{4} disabled for wall-clock cost (\S\ref{sec:composite}) and best-of-$2$ (\alg{6}) removed (\S\ref{sec:ablation}). Each instance is solved once (re-touches return the cached result). The harness, sample, step budget, and grader are held fixed across all eight cells, so the only varying factors are the base model and the stack.

\section{Results and Discussion}
\label{sec:results}

Two results stand out. First, base capability is a large, confound-free effect: single-pass baselines on the identical harness scale cleanly---\textsc{Gemma} $18<$ \textsc{Qwen} $24<$ \textsc{Gpt-mini} $25<$ \textsc{Gpt} $28$ (Table~\ref{tab:4x2})---and the ordering is preserved with the full stack ($22,25,29,34$). Second, the full stack improves every base (on$-$off $+1$ to $+6$), and where we deconfounded it with a no-op control on two strong models the suite's contribution is $+5$ (\textsc{Gpt}) and $+4$ (\textsc{Glm}~5.2)---attributable to the algorithms, not to scaffolding (\S\ref{sec:ablation}); the best configuration is \textsc{Gpt}$+$\emph{Algorithms--A} at $34/52$ ($65\%$).

\begin{table}[t]
\centering
\caption{Model $\times$ stack on a fixed $52$-instance SWE-bench Verified subset (identical harness, official execution-based grader; resolved counts). The full stack improves every base. For the first four models ``off'' is the single-pass baseline, so the on$-$off $\Delta$ also includes a small scaffolding/directive effect (a no-op control on \textsc{Gpt} isolates the algorithm contribution at $+5$, \S\ref{sec:ablation}); for \textsc{Glm}~5.2 the ``off'' cell \emph{is} that no-op control (its single-pass baseline was not run), so its $+4$ is already scaffolding-free.}
\label{tab:4x2}
\small
\begin{tabular}{@{}lccc@{}}
\toprule
Base model & off & \emph{Alg.--A} & $\Delta$ (on$-$off) \\
\midrule
\textsc{Gemma} (\code{gemma4:31b-mlx}, $\sim$31B q.) & $18$ & $22$ & $+4$ \\
\textsc{Qwen} (\code{qwen3.6:27b}, $\sim$27B)        & $24$ & $25$ & $+1$ \\
\textsc{Gpt-mini} (\code{gpt-5.4-mini})              & $25$ & $29$ & $+4$ \\
\textsc{Gpt} (\code{gpt-5.5})                        & $28$ & $\mathbf{34}$ & $+6$ \\
\textsc{Glm} (GLM~5.2)\,$^{\star}$                   & $24$ & $28$ & $+4$ \\
\bottomrule
\end{tabular}

{\centering\footnotesize $^{\star}$\,\textsc{Glm}~5.2 ``off'' $=$ no-op control (scaffolding on, algorithms off).\par}
\end{table}

\begin{figure}[t]
\centering
\begin{tikzpicture}[x=1cm, y=0.135cm,
  bn/.style={fill=black!20, draw=black!55},
  ba/.style={fill=green!25, draw=green!45!black},
  val/.style={font=\scriptsize, inner sep=1pt},
  dl/.style={font=\scriptsize, text=black!55, inner sep=1pt},
  ax/.style={black!70}]
\foreach \y in {0,10,20,30,40} {\draw[black!12] (0,\y) -- (13.5,\y); \draw[ax] (-0.12,\y) -- (0,\y) node[left, font=\scriptsize, xshift=-1pt] {\y};}
\draw[red!55!black, dashed] (0,26) -- (13.5,26) node[right, font=\scriptsize, text=red!55!black] {50\%};
\draw[ax,->] (0,0) -- (0,42);
\draw[ax,->] (0,0) -- (13.9,0);
\foreach \c/\off/\on/\nm/\lf in {1.5/18/22/{\textsc{Gemma}}/{+4}, 4.3/24/25/{\textsc{Qwen}}/{+1}, 7.1/25/29/{\textsc{Gpt-mini}}/{+4}, 9.9/28/34/{\textsc{Gpt}}/{+6}, 12.6/24/28/{\textsc{Glm}}/{+4}} {
  \draw[bn] (\c-0.62,0) rectangle (\c-0.06,\off);
  \draw[ba] (\c+0.06,0) rectangle (\c+0.62,\on);
  \node[val, above] at (\c-0.34,\off) {\off};
  \node[val, above] at (\c+0.34,\on) {\on};
  \node[dl, above=3.4mm] at (\c+0.34,\on) {\lf};
  \node[font=\scriptsize] at (\c,-3.0) {\nm};
}
\node[font=\scriptsize, rotate=90] at (-1.15,21) {resolved\,/\,52};
\draw[bn] (0.25,39) rectangle (0.62,40.4);
\node[anchor=west, font=\scriptsize] at (0.7,39.7) {off};
\draw[ba] (3.0,39) rectangle (3.37,40.4);
\node[anchor=west, font=\scriptsize] at (3.45,39.7) {\emph{Algorithms--A}};
\end{tikzpicture}
\caption{\textbf{Across models.} Resolved instances (of $52$) per base model with the stack off (grey) and on (green); models ordered by baseline. Absolute resolution scales with base capability in both conditions. The on$-$off difference (annotated $+N$) includes a small scaffolding/directive effect, except for \textsc{Glm}~5.2 whose ``off'' bar is the no-op control (\S\ref{sec:ablation}), making its $+4$ already scaffolding-free.}
\label{fig:grid}
\end{figure}
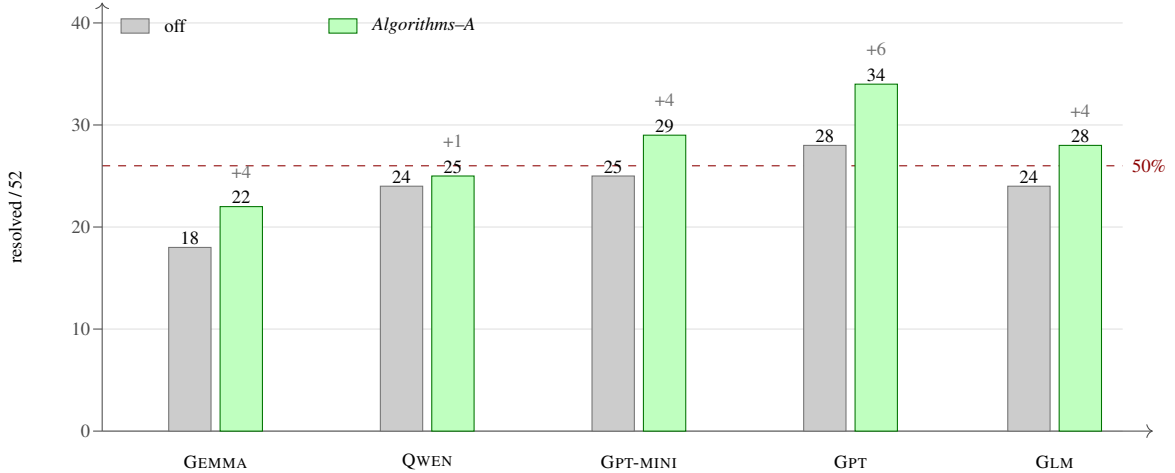

\subsection{Deconfounded single-base ablation}
\label{sec:ablation}

To separate the algorithms from the scaffolding they run inside, we ran a deliberate \emph{no-op composite} control on \textsc{Gpt}---the full composite scaffolding with strategy directives on but every algorithm disabled. It scores $29$ against the single-pass baseline $28$, so the directive/scaffolding effect is only $+1$; the full suite reaches $34$, a \textbf{$+5$ gain over the proper control, attributable to the algorithms themselves}. A second strong model, \textsc{Glm}~5.2, gives the same picture (Table~\ref{tab:4x2}, Fig.~\ref{fig:grid}): $24$ with the algorithms off and $28$ with them on, a $+4$ gain. The deconfounded contribution of the suite is thus positive on both models we controlled ($+5$ and $+4$)---a single run per cell, but a consistent direction across two models.

\begin{figure}[t]
\centering
\begin{tikzpicture}[x=0.26cm, y=0.52cm,
  base/.style={fill=black!18, draw=black!55},
  algc/.style={fill=blue!14, draw=blue!45!black},
  good/.style={fill=green!25, draw=green!45!black},
  bad/.style={fill=red!18, draw=red!55!black},
  lab/.style={font=\scriptsize, inner sep=1pt},
  ax/.style={black!70}]
\foreach \x in {0,10,20,30} {\draw[black!12] (\x,-0.5) -- (\x,12.7); \draw[ax] (\x,-0.5) -- (\x,-0.7) node[below,font=\scriptsize]{\x};}
\draw[red!60!black, dashed] (29,-0.5) -- (29,13.1) node[above,font=\scriptsize,text=red!60!black]{control $29$};
\foreach \y/\val/\st/\nm in {12/28/base/{baseline}, 11/29/base/{control}, 10/30/algc/{\alg{1}}, 9/34/good/{\alg{2}}, 8/31/algc/{\alg{3}}, 7/36/bad/{\alg{4}}, 6/28/algc/{\alg{5}}, 5/26/bad/{\alg{6}}, 4/32/algc/{\alg{7}}, 3/32/algc/{\alg{8}}, 2/29/algc/{\alg{9}}, 1/30/algc/{\alg{10}}, 0/34/good/{full}} {
  \draw[\st] (0,\y-0.3) rectangle (\val,\y+0.3);
  \node[lab, left] at (-0.3,\y) {\nm};
  \node[lab, right] at (\val,\y) {\val};
}
\node[font=\scriptsize] at (18,-1.5) {resolved\,/\,52};
\end{tikzpicture}
\caption{\textbf{Single-base ablation (\textsc{Gpt}, \code{gpt-5.5}).} Resolved instances for each single-algorithm config, the no-op composite \emph{control} ($29$, dashed), the single-pass baseline ($28$), and the full suite ($34$); one run per cell (Table~\ref{tab:ablation}). Most configs sit near the control; the full suite and \alg{2} are highest (green). \alg{4}'s $36$ is an artifact---its gate accepted $0$ edits, so it ran the control---and \alg{6} (best-of-$2$, removed) is below the control (both red). Single runs on expensive evaluations; per-cell deltas are not separated from run-to-run variance.}
\label{fig:ablation}
\end{figure}
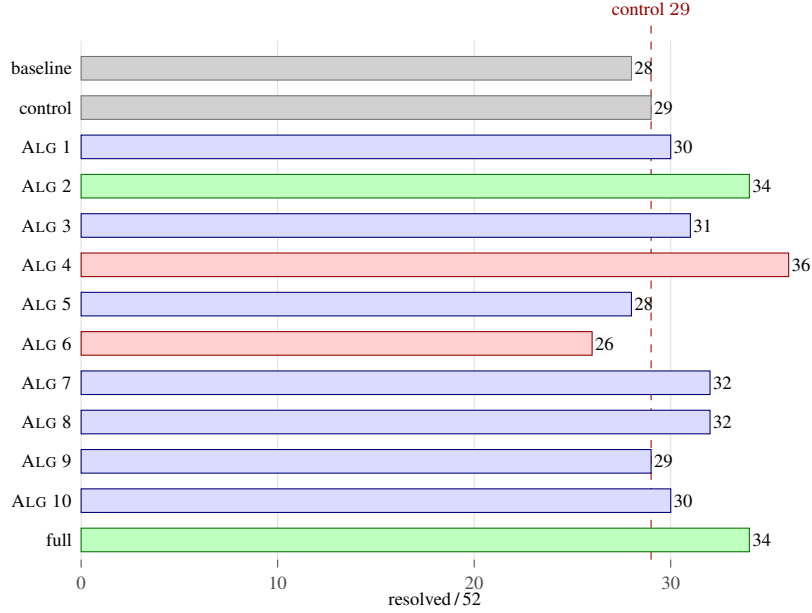

Decomposing the \textsc{Gpt} suite into single-algorithm runs (Table~\ref{tab:ablation}, Fig.~\ref{fig:ablation}) localizes the contribution. Re-anchored to the $29$ control, several algorithms contribute alone: \alg{2} $+5$ (the strategy-directive learner), \alg{7}/\alg{8} $+3$ (micro-step search, self-oracles), \alg{3} $+2$; the full suite reaches the best single component, consistent with overlapping rather than additive gains. Two honest reads: \alg{6} (best-of-$2$) is net-negative ($26$) and, per the run logs, never produced a second attempt while adding patch-apply failures, so we \emph{remove} it from the live stack; and \alg{4}'s $36$ is \emph{not} an algorithm effect---its confidence-sequence gate accepted zero edits, so it ran the control configuration and the count is a high draw, which the event-log attribution (not the raw count) caught. These are single-run figures (\S\ref{sec:limitations} collects the limitations and next steps).

\begin{table}[t]
\centering
\caption{\textsc{Gpt} (\code{gpt-5.5}) single-algorithm ablation, one run per cell, same $52$ instances and grader. The no-op composite (directives on, all algorithms off) is the deconfounded control; the full suite reaches $34$, $+5$ over it. \alg{4}'s $36$ is not an algorithm effect---its gate accepted $0$ edits, so it ran the control configuration (caught by the event log).}
\label{tab:ablation}
\small
\begin{tabular}{@{}lcl@{}}
\toprule
Config & resolved\,/\,52 & note \\
\midrule
single-pass baseline      & $28$ & no strategy directive \\
no-op composite (control) & $29$ & directives on, algorithms off \\
\alg{1} & $30$ & \\
\alg{2} & $34$ & directive learner ($+5$ over control) \\
\alg{3} & $31$ & \\
\alg{4} & $36$ & $0$ accepts --- ran control config \\
\alg{5} & $28$ & $\approx 0$ accepts \\
\alg{6} & $26$ & best-of-$2$; never fired, apply failures --- removed \\
\alg{7} & $32$ & micro-step search \\
\alg{8} & $32$ & self-oracles \\
\alg{9} & $29$ & \\
\alg{10} & $30$ & \\
full suite & $\mathbf{34}$ & $+5$ over control \\
\bottomrule
\end{tabular}
\end{table}

\subsection{What the mechanisms verifiably do}
\label{sec:mechanism}

The variance-immune evidence is in the event logs (decisions, accepts, oracle admissions, veto and react/refine firings), which pass-count noise does not touch. Log analysis suggests the measurable lever inside the stack is \emph{more closed-loop debugging}, driven by the self-oracle (\alg{8}) and reflected in the full suite: relative to the control, \code{run\_tests} calls rise by $\sim$50\% and mean episode length by $\sim$1.3 steps---the agent iterates against its self-authored tests rather than guessing once. Among the certificate controllers, the ones that fire every round are the directive learners \alg{2} and \alg{3} ($52$ policy updates each; \alg{2} is the top single-algorithm cell at $34$); the gated, library, and forgetting controllers \alg{4}, \alg{5}, \alg{1} contribute less. On this evidence the suite's live value is carried by \alg{2}/\alg{3} (directive shaping) and \alg{7}/\alg{8} (verified search and self-oracle). Consistent with selection-not-creation, the stack does not destabilize a base---added regressions are few (full suite $4$ p2p-flagged instances vs.\ $3$ for the control)---and a consensus-flip analysis separates signal from noise: a robust subset of \textsc{Gpt} instances flips under several distinct configs while single-config flips are noise-like; $17/52$ are solved by every config and $10$ by none.

\textbf{Position.} Two levers move the metric, and they are not interchangeable. Within a fixed base the lever is the algorithm suite, whose deconfounded gain is carried by \alg{2}/\alg{3} (directive shaping) and \alg{7}/\alg{8} (verified search and self-oracle). For \emph{absolute} resolution the lever is a stronger base: the $10$ instances solved by no configuration---mostly Matplotlib---are a capability/harness wall that more search does not move.

\section{Limitations}
\label{sec:limitations}

The endogenous-loop guarantees remain open conjectures: each controller's statistical primitives are individually published and sound, but their compositions---the two-timescale coupling in \alg{2}, the backward-transfer reduction in \alg{1}, the performative corrections in \alg{1} and \alg{4}---are not proven here, and several rest on assumptions the endogenous loop itself erodes (a stationary held-out real set for \alg{2}; a learnable fully-predictive augmented outcome and bounded comparator path variation for \alg{3}; exchangeable task generation for \alg{5}). The performative sensitivity $\varepsilon$ is a behavioral constant taken as a hyperparameter; it is not estimable online with its own validity guarantee, and the contraction condition $\varepsilon L<1$ may fail at LLM scale. The anytime-validity that makes the gates sound also makes them conservative: PAC-Bayes terms grow vacuous for high-dimensional posteriors, the DV forgetting floor rises with $t$, and \alg{4} guarantees safety, not progress. A boundary we state up front: controllers and verifier-guided search select, gate, and reshape the behavior of the frozen base; they cannot manufacture capability it lacks. This bounds the absolute \emph{ceiling}, which the base sets; within that ceiling the suite's gains (\S\ref{sec:ablation},~\S\ref{sec:mechanism}) are real but measured as single runs on expensive evaluations, so we report magnitudes rather than significance. The self-oracle is fallible by design---it can under-claim (an over-strict oracle fails to flip a correct patch) and over-claim (a false-green pass)---so a patch is reported as ``promising,'' never ``resolved,'' the terminal grader being authoritative. A proper ablation across diverse harnesses and base models is also outstanding and, we think, the most informative next step: with one harness and a single sample per cell we can say the suite as a whole helps, but not which algorithm is advantageous for which \emph{type} of problem---the matching of controllers to problem structure (stochastic vs.\ systematic failures, short vs.\ long horizons, repository families) can only be settled by a factorial study that varies harness and model together. Finally, each cell is a single run (no per-controller multi-seed isolation) and the slow-loop distillation is designed but not trained; repeated runs, a stronger-base confirmation, the per-task algorithm--problem ablation, and the distillation step are the remaining experiments.

\section{Conclusion}
\label{sec:conclusion}

SEA confines self-evolution of an LLM agent to a frozen base plus a small steering adapter and a versioned harness, and admits each self-modification through an anytime-valid gate that certifies it against a budgeted error ledger. Five controllers compose published guarantees, and five verifier-in-the-loop mechanisms---including a self-authored reproduction-oracle verifier computed from the issue alone---supply the dense, grader-free signal those gates need. On a $52$-instance SWE-bench Verified subset across four bases, base capability is the dominant, confound-free effect, and on two strong base models a deliberate no-op-composite control isolates the suite's contribution at $+5$ and $+4$ (\textsc{Gpt} $29\to34$, $65\%$; \textsc{Glm}~5.2 $24\to28$), with the event logs verifying that its mechanisms fire and prevent regressions. Because these single-run evaluations are expensive, confirming run-to-run variance and adapting the per-task algorithm mix---then the distillation step that turns verified traces into weights---are the outstanding work.

\bibliographystyle{plainnat}
\bibliography{references}

\appendix

\section{Algorithm Pseudocode}
\label{app:pseudocode}

This appendix collects the pseudocode for every algorithm in Table~\ref{tab:algmap}; each box is referenced from the main text.

\begin{algorithm}[t]
\caption{PPB-CL: forgetting-gated, trust-regioned continual adapter learning}
\label{alg:ppbcl}
\begin{algorithmic}[1]
\State \textbf{init} frozen prior $Q_0=\mathcal{N}(\theta_0,\mathrm{diag}\,e^{v})$; update-direction buffer $B\gets\emptyset$ (FIFO, cap $m$); anchors $\mathcal{A}\gets\emptyset$ (cap $A$, each task with best historical reward)
\For{$t=1,2,\dots$}
  \State rollouts $\gets\textsc{Deploy}(\pol_{t-1}, n_t \text{ tasks})$; record rewards $R_i$, directive indices $a_i$ \Comment{$S_t\sim\dist(\pol_{t-1})$}
  \State $\hat g\gets\frac{1}{n_t}\sum_i\big(\ell_i-\bar\ell_{\mathrm{pool}}\big)\,\nabla_\theta\log p_\theta(a_i)$,\quad $\ell_i=1-R_i$ \Comment{baseline pooled over the recent replay window}
  \State $\hat g_\perp\gets\hat g-\sum_{J\in B}\frac{\langle\hat g,J\rangle}{\langle J,J\rangle}J$ \Comment{OGD projection off stored directions}
  \State $\theta_{\mathrm{cand}}\gets\theta_{t-1}-\eta\,\hat g_\perp$;\quad $B_{\mathrm{fgt}}\gets\textsc{DVBound}(\theta_{\mathrm{cand}},t)$ \Comment{Eq.~\eqref{eq:dv}}
  \State $d\gets0$
  \While{$B_{\mathrm{fgt}}>\tau_{\mathrm{forget}}$ \textbf{and} $d<5$} \Comment{damped re-test}
     \State $\theta_{\mathrm{cand}}\gets\theta_{t-1}+\rho\,(\theta_{\mathrm{cand}}-\theta_{t-1})$;\quad $B_{\mathrm{fgt}}\gets\textsc{DVBound}(\theta_{\mathrm{cand}},t)$;\quad $d\gets d+1$
  \EndWhile
  \State $r\gets\max(\tau_{\mathrm{forget}}-B_{\mathrm{fgt}},\,0)/\varepsilon$ \Comment{performative trust region}
  \If{$\lVert\theta_{\mathrm{cand}}-\theta_{t-1}\rVert_2>r$}
     \State $\theta_{\mathrm{cand}}\gets\Pi_{\mathcal{B}(\theta_{t-1},r)}(\theta_{\mathrm{cand}})$; recompute $B_{\mathrm{fgt}}$ and $r$
  \EndIf
  \If{$B_{\mathrm{fgt}}\le\tau_{\mathrm{forget}}$ \textbf{and} $\varepsilon L<1$} \Comment{\textsc{accept}}
     \State $\theta_t\gets\theta_{\mathrm{cand}}$;\quad $B\gets(B\cup\{\hat g_\perp\})[-m{:}]$;\quad update $\mathcal{A}$ with this round's tasks/best rewards
  \Else\ $\theta_t\gets\theta_{t-1}$ \Comment{\textsc{hold}}
  \EndIf
  \State emit certificate $(t,\,B_{\mathrm{fgt}},\,r,\,\KL(Q_{\theta_t}\|Q_0),\,\text{PAC-Bayes penalty},\,d)$
\EndFor
\Statex
\Function{DVBound}{$\theta$, $t$}\Comment{anytime-valid backward transfer, Eq.~\eqref{eq:dv}}
  \State deploy $\pol(\theta)$ on anchor tasks; $\widehat{\mathrm{bt}}\gets\frac{1}{|\mathcal{A}|}\sum_{j}\big[(1-R_j)-(1-R^*_j)\big]$
  \State \Return $\widehat{\mathrm{bt}}+A_t/\lambda^*+\lambda^*/(8|\mathcal{A}|)$ \Comment{$A_t=\KL+\log(2\sqrt t/\delta)$; $\lambda^*=\sqrt{8|\mathcal{A}|A_t}$, the DV-optimal temperature}
\EndFunction
\end{algorithmic}
\end{algorithm}

\begin{algorithm}[t]
\caption{PNMP-A: two-timescale anchored preference learning with an e-value drift gate}
\label{alg:pnmpa}
\begin{algorithmic}[1]
\State \textbf{init} $q\gets\mathbf{0}$; frozen anchor $q_{\mathrm{real}}$; magnet $z\gets\theta_0$; reference $\log\pol_{\mathrm{ref}}\gets\log\mathrm{softmax}(\theta_0)$; CTHS budget($\delta_0$); rates $a^{\mathrm{slow}}_t=\tfrac12(t{+}1)^{-1}$, $a^{\mathrm{fast}}_t=\tfrac12(t{+}1)^{-0.7}$
\For{$t=1,2,\dots$}
  \State rollouts $\gets\textsc{Deploy}(\pol_{t-1})$; pairs $\gets$ \{$(i,j,\text{winner})$ : winner has higher mean reward over the recent replay window\}
  \If{\textbf{not} budget.exhausted} \Comment{\textsc{slow}: anchored preference update}
    \State $g_{\mathrm{synth}}\gets\nabla_q\,\widehat{\mathcal{L}}_{\mathrm{BT}}(q;\text{pairs})$;\quad $q_{\mathrm{cand}}\gets q+a^{\mathrm{slow}}_t\big[\alpha\,(q_{\mathrm{real}}-q)+(1-\alpha)\,g_{\mathrm{synth}}\big]$
    \State $X\gets\big\{\,\lvert\sigma(q^{\mathrm{cand}}_i{-}q^{\mathrm{cand}}_j)-\sigma(q^{\mathrm{real}}_i{-}q^{\mathrm{real}}_j)\rvert\,\big\}_{i<j}$ \Comment{per-pair drift vs.\ frozen anchor}
    \State $E\gets\textsc{EProcess}(X;\,H_0\!:\E[X]\le\tau)$
    \State $k\gets k+1$;\quad $\delta_k\gets\delta_0/\big(Z\,k\log^2(k{+}1)\big)$ \Comment{normalized CTHS spend}
    \If{$E\ge1/\delta_k$} reject: keep $q$ \Comment{anytime-valid evidence of unsafe drift}
    \Else\ $q\gets q_{\mathrm{cand}}$ \Comment{\textsc{accept}}
    \EndIf
  \EndIf
  \State $\mathrm{adv}_i(p)\gets\sum_j p_j\,\sigma(q_i-q_j)$;\quad $\mathrm{prox}(g)\gets\mathrm{softmax}\!\big(\tfrac{\log p_{t-1}+\eta g+\beta\log\pol_{\mathrm{ref}}}{1+\beta}\big)$ \Comment{\textsc{fast}}
  \State $\theta_{1/2}\gets\log\mathrm{prox}\big(\mathrm{adv}(\mathrm{softmax}(\theta_{t-1}))\big)$;\quad $\theta_{\mathrm{MP}}\gets\log\mathrm{prox}\big(\mathrm{adv}(\mathrm{softmax}(\theta_{1/2}))\big)$
  \State $\theta_t\gets\theta_{t-1}+a^{\mathrm{fast}}_t\,(\theta_{\mathrm{MP}}-\theta_{t-1})$ \Comment{extragradient step}
  \If{$t\bmod K=0$} $z\gets\theta_t$ \Comment{magnet refresh}
  \EndIf
  \State emit certificate $(t,\,\textsc{accept}/\textsc{hold},\,\delta_k,\,E,\,\KL(\pol_t\|z),\,\text{pref.\ drift})$
\EndFor
\end{algorithmic}
\end{algorithm}

\begin{algorithm}[t]
\caption{PA-COCOA: counterfactual credit assignment driving a coin-betting oracle}
\label{alg:pacocoa}
\begin{algorithmic}[1]
\State \textbf{init} oracle $\mathcal{O}\gets$ per-coordinate KT coin betting (dim $k$); $\theta_0\gets\mathcal{O}.\textsc{Predict}()$; reward stream $V\gets[\,]$; $V_T\gets0$
\For{$t=1,2,\dots$}
  \State rollouts $\gets\textsc{Deploy}(\pol_t)$ \Comment{actor samples directive $a\sim\mathrm{softmax}(\theta_t)$ per task}
  \State append masked mean process reward to $V$ (environment-errored rollouts zeroed); extend replay buffer \Comment{augmented outcome: rollouts carry $(a,\pol_t)$}
  \For{each directive $a\in\{1..k\}$} \Comment{contribution model, off-policy on replay}
     \State $\widehat{w}(a)\gets\dfrac{\sum_{r\in\mathrm{recent}(4h):\,a_r=a}\omega_r\,2^{-\mathrm{age}(r)/h}\,\tilde R_r}{\sum_{r:\,a_r=a}\omega_r\,2^{-\mathrm{age}(r)/h}}$\quad (default $\tfrac12$ if no data) \Comment{$\tilde R$: process reward, Eq.~\eqref{eq:process}; $\omega_r$ folds in the env-error mask and repo-family weight, and rejected-candidate rollouts are skipped}
  \EndFor
  \State $p\gets\mathrm{softmax}(\theta_t)$;\quad $g\gets(\widehat{w}\odot p)-p\,\langle\widehat{w},p\rangle+\varphi\,(\mathbf{1}/k-p)$ \Comment{Eq.~\eqref{eq:cocoa-grad} $+$ exploration floor; no importance ratio}
  \State $\mathrm{drift}\gets\textsc{WildBootstrapTrend}(V;\,\alpha{=}0.01)$ if $|V|\ge12$ else \textbf{false}
  \State \textbf{if} drift \textbf{then} $\mathcal{O}.\textsc{Restart}()$ \Comment{anchor a new comparator segment}
  \State $\theta_{t+1}\gets\mathcal{O}.\textsc{Update}(-g)$ \Comment{oracle minimizes loss; feed $-g$}
  \State $V_T\gets V_T+\lVert\theta_{t+1}-\theta_t\rVert_2$
  \State emit certificate $(t,\,\lVert g\rVert,\,\lVert\theta_{t+1}-\theta_t\rVert,\,V_T,\,\#\text{restarts},\,\max_a\widehat w-\min_a\widehat w)$
\EndFor
\end{algorithmic}
\end{algorithm}

\begin{algorithm}[t]
\caption{SGM-CS: self-edit admission with anytime familywise risk control}
\label{alg:sgmcs}
\begin{algorithmic}[1]
\State \textbf{init} deployed policy $\pol_0$; confirmation count $k\gets0$; CTHS budget($\delta_0$); per-version CS map
\For{$t=1,2,\dots$}
  \State \textbf{if} budget.exhausted \textbf{then} emit \textsc{hold}; \textbf{continue} \Comment{harness frozen}
  \State $\pol_{\mathrm{cand}} \gets$ pop a requeued promising \emph{same-family} harness (its repo among the round's), else $\textsc{ProposeEdit}(\pol_{t-1}, t)$ \Comment{shadow requeue / force-edit / budget / guidance}
  \State \textbf{if} $\mathrm{version}(\pol_{\mathrm{cand}})=\mathrm{version}(\pol_{t-1})$ \textbf{then} emit \textsc{hold}; \textbf{continue} \Comment{no-op edit; no spend}
  \State $\text{tasks} \gets$ rank batch by \emph{weak recent evidence} (unseen / low-reward first); move the \emph{best-evidenced} task to the front as pilot \Comment{higher-signal, index-aligned pairing}
  \State \textbf{pre-gate:} run $\pol_{\mathrm{cand}}$ on the pilot; \textbf{if} best pilot reward $<$ \textsc{pregateMin} \textbf{and} pilot task is \emph{known passable} \textbf{then} spend $\delta_k$ ($k{+}{+}$); requeue a promoted shadow; emit \NSF{}; \textbf{continue} \Comment{evidence-gated; cold start falls through}
  \State deploy $\pol_{t-1}$ and $\pol_{\mathrm{cand}}$ concurrently on the \emph{same} task batch (isolated working copies; common random numbers---aligned per-task seeds)
  \State \textbf{if} no learnable pairing survives (e.g.\ every rollout env-errored) \textbf{then} emit \textsc{hold}; \textbf{continue} \Comment{no spend}
  \State $k\gets k+1$;\quad $\delta_k \gets \delta_0 / \big(Z\,k\log^2(k+1)\big)$ \Comment{normalized CTHS spend, committed on decision}
  \State update per-version CSs \citep{howard2021time} with $R_{\mathrm{base}}$, $R_{\mathrm{cand}}$
  \State $\mathrm{LCB} \gets \mathrm{CS}_{\delta_k}^{\mathrm{lower}}\big(R_{\mathrm{cand}}-R_{\mathrm{base}}\big) - \varepsilon\cdot\Wone(R_{\mathrm{cand}},R_{\mathrm{base}})$ \Comment{performative correction, exact 1-D $\Wone$}
  \If{wild-bootstrap trend test rejects stationarity of the $v_{\mathrm{base}}$ stream from baseline-evaluated rounds}
     \State $\mathrm{LCB}\gets\mathrm{LCB}-\mathrm{radius}$ \Comment{widen; \citet{chandak2020safe}}
  \EndIf
  \If{$\mathrm{LCB} \ge -\epsilon_{\mathrm{tol}}$} \State $\pol_t\gets\pol_{\mathrm{cand}}$ \Comment{\textsc{accept}}
  \Else\ $\pol_t\gets\pol_{t-1}$; \textbf{if} $\pol_{\mathrm{cand}}$ produced a strong patch \textbf{then} requeue it (bounded retries per harness) \Comment{\NSF{}, but retry a promising harness later}
  \EndIf
  \State emit certificate $(t,\ \textsc{accept}/\NSF,\ \mathrm{LCB},\ \delta_k,\ \varepsilon\Wone,\ \text{widened?},\ \text{shadow})$
\EndFor
\end{algorithmic}
\end{algorithm}

\begin{algorithm}[t]
\caption{SDC-QD: library growth with MDL compression and quality-diversity acceptance}
\label{alg:sdcqd}
\begin{algorithmic}[1]
\State \textbf{init} corpus $C\gets\emptyset$; MAP-Elites archive $\mathcal{M}$ over descriptor $\phi$; library $L\gets L_0$; solve threshold $s^*$
\For{$t=1,2,\dots$}
  \For{each task in batch} \Comment{\textsc{wake}: LLM-guided program search}
     \State $\rho\gets\textsc{ParseSExpr}\big(\mathrm{LLM}(\text{task},\,L_t)\big)$ \Comment{untrusted output; \textbf{None} if not well-formed}
     \State \textbf{if} $\rho\ne$ \textbf{None and} $\mathrm{env.score}(\rho)\ge s^*$ \textbf{then} $C\gets C\cup\{\rho\}$
  \EndFor
  \State $J_{\mathrm{before}}\gets|L_t|+\sum_{\rho\in C}|\rho|$
  \State cands $\gets$ subtrees$(C)\;\cup$ pairwise antiunifications of compatible subtrees \Comment{\textsc{sleep}: Stitch}
  \State $A^*\gets\arg\max_{A\in\text{cands}}u(A)$,\quad $u(A)=m_A\,(b_A-1-a_A)-b_A$ \Comment{descending scan, sound dominance break}
  \If{$A^*\ne$ \textbf{None and} $\Delta J\coloneqq u(A^*)\ge u_{\min}$} \Comment{quality-diversity acceptance, minimum-utility bar}
     \State $b\gets\phi(L_t\cup\{A^*\})=(\,|L_t|{+}1,\ \mathrm{mean}_{\rho\in C}|\rho|\,)$;\quad $f\gets J_{\mathrm{before}}-\Delta J$ \Comment{lower fitness is better}
     \If{$\mathcal{M}.\textsc{IsNovel}(b,f)$} \Comment{empty cell or improves incumbent}
        \State $\mathcal{M}.\textsc{Add}(L_t\cup\{A^*\},\,b,\,f)$;\quad $L_{t+1}\gets L_t\cup\{A^*\}$; surface $A^*$ in harness $L_2$ \Comment{\textsc{accept}}
     \Else\ $L_{t+1}\gets L_t$ \Comment{\textsc{reject}: no novelty}
     \EndIf
  \Else\ $L_{t+1}\gets L_t$ \Comment{\textsc{reject}: no MDL gain}
  \EndIf
  \State cert $\gets$ description-length certificate, Eq.~\eqref{eq:catoni}
  \State emit certificate $(t,\,\#\text{solved},\,|L_{t+1}|,\,\Delta J,\,\mathcal{M}.\mathrm{coverage},\,|\mathcal{M}.\mathrm{ParetoFrontier}()|,\,\text{cert})$
\EndFor
\end{algorithmic}
\end{algorithm}

\begin{algorithm}[t]
\caption{(\alg{6}) Verifier-in-the-loop search: best-of-$N$ selection and refinement with backtracking}
\label{alg:verifier}
\begin{algorithmic}[1]
\Function{BestOfN}{$\pol$, task $x$;\ $n$} \Comment{independent diverse attempts; verifier selects}
  \For{$i=0,\dots,n-1$} $y_i\gets\textsc{Act}(\pol,x,\,\mathrm{attempt}{=}i)$ \Comment{attempt-indexed seed + isolated workspace; live: $n{=}2$, attempt $2$ only on a failing verdict, prompted to differ structurally}
  \EndFor
  \State \Return $y_{i^*}$,\quad $i^*=\arg\max_i V(x,y_i)$ \Comment{$V$: shaped native-test verifier; $p\to1-(1-p)^n$}
\EndFunction
\Statex
\Function{Refine}{$\pol$, task $x$;\ depth $d$} \Comment{verifier-guided hill-climb over patches}
  \State $y^*\gets\textsc{Act}(\pol,x,\,\mathrm{attempt}{=}0)$;\quad $v^*\gets V(x,y^*)$
  \For{$i=1,\dots,d-1$}
     \State \textbf{if} $v^*\ge1$ \textbf{then break} \Comment{resolved; stop early}
     \State $x'\gets x\oplus\big(y^*,\ \textsc{Feedback}(x,y^*)\big)$ \Comment{prior best patch + why the verifier rejected it}
     \State $y\gets\textsc{Act}(\pol,x',\,\mathrm{attempt}{=}i)$;\quad $v\gets V(x,y)$ \Comment{scored on the \emph{original} task}
     \State \textbf{if} $v>v^*$ \textbf{then} $(y^*,v^*)\gets(y,v)$ \Comment{climb; else backtrack (keep best)}
  \EndFor
  \State \Return $y^*$
\EndFunction
\end{algorithmic}
\end{algorithm}

\begin{algorithm}[t]
\caption{(\alg{7}) Verified micro-step search (fast loop): beam search with memory and a verifier cascade}
\label{alg:micro}
\begin{algorithmic}[1]
\State \textbf{init} beam $\gets\{$seed patch$\}$; memory $\mathcal{M}\gets\emptyset$; best $\gets$ seed; depth $\gets 0$
\While{depth $<$ \textsc{maxDepth} \textbf{and} $V(\text{best})<1$}
  \State children $\gets[\,]$
  \For{node in beam}
     \State $H\gets\textsc{Generate}(\text{node},\,\mathcal{M},\,k)$ \Comment{reasoning-first: name class/method, then $k$ distinct one-line edits; capped prompt (issue $\le$1.8k chars, JSON-array reply)}
     \For{$h\in H$ with $h\notin\mathcal{M}.\text{tried}$}
        \State $\rho\gets\textsc{Compose}(\text{node.patch},\,h)$;\quad \textbf{if} $\rho=\bot$ \textbf{then} $\mathcal{M}.\text{tried}\!\mathrel{+}=h$; \textbf{continue} \Comment{did not apply}
        \State $(v,\,\text{fb})\gets\textsc{Cascade}(\rho)$ \Comment{cheap parse/apply check, then native suite only on survivors}
        \State $\mathcal{M}.\textsc{Record}(h,v,\text{fb})$;\quad children.append$(\rho,v)$;\quad \textbf{if} $v\ge1$ \textbf{then break}
     \EndFor
  \EndFor
  \State depth $\gets$ depth $+1$;\quad \textbf{if} children $=\emptyset$ \textbf{then break} \Comment{a whole diverse round moved nothing}
  \State beam $\gets$ top-$b$ of (beam $\cup$ children) by $V$;\quad best $\gets\arg\max_V$ over all scored candidates \Comment{parents kept; best tracked globally, independent of beam policy}
\EndWhile
\State \Return best, its verified trace, and \#\,expensive verifier calls
\end{algorithmic}
\end{algorithm}

\begin{algorithm}[t]
\caption{(\alg{8}) Self-authored reproduction oracles: grader-free in-loop verification}
\label{alg:firewall}
\begin{algorithmic}[1]
\State \textbf{given} issue text (no gold tests, no test patch, no f2p/p2p lists); a verifier model; a solver agent
\State $O \gets \textsc{Synthesize}(\text{issue},\,k)$ over an \emph{ensemble} of models \Comment{any model may author a usable oracle (recall)}
\State $A \gets \{\,o\in O : \textsc{RunOnBase}(o)\text{ fails, not a timeout, not a \emph{syntactic} self-error}\,\}$ \Comment{fails-on-base admission---no ground truth}
\State $A \gets \{\,o\in A : \textsc{SymptomJudge}(o)\text{ matches the issue}\,\}$ \Comment{denoise; separates real import bugs from hallucinated APIs}
\State the agent debugs with \code{run\_tests} $\!=\!$ self-score over $A$: \Comment{the in-loop gradient is $A$, never the grader}
\Statex \quad $V_{\mathrm{self}}(\rho) = \dfrac{\#\{o\in A:\ \rho\text{ flips }o\text{ fail}\to\text{pass}\}}{|A|}$, set to $0$ if any green check regresses; \emph{promising} iff all flip
\State $\rho^\star \gets$ the agent's submitted patch
\State \textbf{measure once} (terminal, held out from the loop): held-out grader on $\rho^\star$ \Comment{used only to report, never to steer}
\end{algorithmic}
\end{algorithm}

\begin{algorithm}[t]
\caption{(\alg{9}) Efficient continual search: search-layer controllers (compute-allocation $+$ QD diversity; step-credit $+$ forgetting gate)}
\label{alg:continual}
\begin{algorithmic}[1]
\State \textbf{init} step-credit $\mathcal{C}$ (\alg{3}); forgetting gate $\mathcal{G}$ (\alg{1}); earlier-repo anchor $A$
\For{each round over the problem stream}
  \For{each problem $x$}
     \State order branch classes for $x$ by learned credit $\mathcal{C}$ \Comment{\alg{3}: productive step macros first}
     \State $\textsc{BeamSearch}(x)$ with two controllers active: \Comment{Algorithm~\ref{alg:micro}}
     \State \quad \alg{4} \emph{budget}: record per-depth marginal gain; stop when the CS upper bound $\le$ \textsc{minGain}
     \State \quad \alg{5} \emph{diversify}: pick the next beam by MAP-Elites behavior cells, not raw top-$b$ score
     \State credit the solving branch $1$, the branches tried before it $0$ \Comment{\alg{3} update; wasted shapes demoted}
  \EndFor
  \State propose policy update; \textbf{accept iff} $\mathcal{G}$: re-eval of anchor $A$ does not regress \Comment{\alg{1} forgetting gate}
  \State record (mean depth, resolved-rate, forgetting) for the round
\EndFor
\end{algorithmic}
\end{algorithm}

\begin{algorithm}[t]
\caption{(\alg{10}) Verified self-repair: adopt a harness repair only by measured fix-rate}
\label{alg:repair}
\begin{algorithmic}[1]
\Function{ProposeRepairs}{failures $F$ (base, hypothesis pairs that did not apply); repertoire $\mathcal{R}$; threshold $\tau$}
  \State adopted $\gets[\,]$
  \For{each primitive $r\in\mathcal{R}$}
     \State fixed $\gets \big|\{(b,h)\in F : r(h)\ne h \ \wedge\ \textsc{Composes}(b,\,r(h))\}\big|$ \Comment{re-test against the \emph{real} composer}
     \State \textbf{if} fixed$/|F| \ge \tau$ \textbf{then} adopted.append$\big(r,\ \text{fixed}/|F|\big)$ \Comment{credited only if it changed $h$ and applied}
  \EndFor
  \State \Return adopted sorted by fix-rate, best-first \Comment{appended to the harness repair pipeline ($L_2$)}
\EndFunction
\Statex
\Function{ProposeGenerationRepair}{suffix repertoire $\mathcal{G}$; baseline apply-rate $\rho_0$; margin $m$}
  \For{each suffix $g\in\mathcal{G}$}
     \State $\rho \gets \textsc{RegenApplyRate}(g)$ \Comment{re-generate hypotheses with $g$ appended; fraction that compose}
     \State \textbf{if} $\rho-\rho_0\ge m$ \textbf{then} adopt $g$ \Comment{verified at the source: fixes \emph{future} candidates}
  \EndFor
\EndFunction
\end{algorithmic}
\end{algorithm}

\begin{algorithm}[t]
\caption{Composite controller: scheduled co-evolution over one shared policy}
\label{alg:composite}
\begin{algorithmic}[1]
\State \textbf{init} shared policy $\pol$; schedule $\{(\text{name}_i,\,\mathcal{C}_i,\,K_i)\}$ \Comment{$K_i=0$ disables $\mathcal{C}_i$ (ablation)}
\For{$t=1,2,\dots$}
  \State ran $\gets[\,]$
  \For{each $(\text{name}_i,\mathcal{C}_i,K_i)$ with $K_i>0$ \textbf{and} $t\bmod K_i=0$} \Comment{in schedule order}
     \State \textbf{if} cost-aware \textbf{and} $\mathcal{C}_i$ is a slow $L_2$ controller with no \emph{committed} gain for the last \textsc{skipAfter} rounds \textbf{then} skip ($\textsc{hold}$, reclaim its deploys); set its counter to $\textsc{skipAfter}-1$ (probe next round); \textbf{continue}
     \State $\mathcal{C}_i.\pol\gets\pol$;\quad $c_i\gets\mathcal{C}_i.\textsc{Step}(t)$;\quad $\pol\gets\mathcal{C}_i.\pol$ \Comment{propagate this layer's edit}
     \State update $\mathcal{C}_i$'s no-gain counter from $c_i$ (reset on committed gain: accept / shadow-accept / committed macro / anti-macros); ran.append$\big((\text{name}_i,c_i)\big)$
  \EndFor
  \State emit merged certificate: metrics prefixed $\text{name}_i.\ast$; spends summed;
  \Statex \hspace{\algorithmicindent}\quad decision $\gets$ first of (\textsc{accept}, \NSF{}, \textsc{reject}) present among $\{c_i\}$, else \textsc{hold}
\EndFor
\end{algorithmic}
\end{algorithm}

\section{The Full Self-Evolution Loop: How \alg{6}--\alg{10} Plug In}
\label{app:fullloop}

\begin{figure}[t]
\centering
\resizebox{\textwidth}{!}{%
\begin{tikzpicture}[font=\footnotesize,
  layer/.style={draw=blue!45!black!60, rounded corners=2pt, minimum width=4.7cm, minimum height=6.6mm, align=center, fill=blue!8, font=\scriptsize},
  slowc/.style={draw=orange!75!black, rounded corners=2pt, text width=4.9cm, inner ysep=2.4pt, align=left, fill=orange!14, font=\scriptsize},
  fastc/.style={draw=green!45!black, rounded corners=2pt, text width=4.9cm, inner ysep=2.4pt, align=left, fill=green!12, font=\scriptsize},
  guardc/.style={draw=violet!70!black, rounded corners=2pt, text width=4.9cm, inner ysep=2.4pt, align=left, fill=violet!10, font=\scriptsize},
  eng/.style={rounded corners=2pt, text width=5.2cm, inner ysep=2.6pt, align=left, font=\scriptsize},
  enga/.style={eng, draw=black!55, fill=black!5},
  engf/.style={eng, draw=green!45!black, fill=green!12},
  engv/.style={eng, draw=violet!70!black, fill=violet!11},
  engs/.style={eng, draw=orange!78!black, fill=orange!14},
  aux/.style={draw=black!70, rounded corners=2pt, align=center, fill=yellow!16, inner ysep=3pt, font=\scriptsize},
  arr/.style={-{Stealth[length=2.2mm]}, semithick},
  lab/.style={font=\scriptsize, fill=white, inner sep=1pt}
]
\node[layer, fill=black!14, draw=black!60] (L0) {$L_0$ --- frozen base model};
\node[layer, above=2.6mm of L0] (L1) {$L_1$ --- steering adapter $\theta$ (online-steered)};
\node[layer, above=2.6mm of L1] (L2) {$L_2$ --- harness: prompt $\cdot$ tools $\cdot$ budgets $\cdot$ library $\cdot$ repair};
\node[draw=blue!45!black!70, dashed, rounded corners=3pt, fit=(L0)(L1)(L2), inner sep=2.6mm,
      label={[font=\scriptsize, text=blue!45!black, align=center]270:{deployed policy\\ $\pol_t=L_0\circ L_1^{(t)}\circ L_2^{(t)}$}}] (POL) {};
\node[enga, right=20mm of L2.east, anchor=north west] (actor) {\textbf{actor} --- multi-step \textsc{ReAct} agent over the structured tool vocabulary (\S\ref{sec:composite})};
\node[engf, below=2.4mm of actor] (a6)  {\alg{6} \textbf{best-of-$N$ / refinement} --- designed operator; off in the live stack};
\node[engf, below=2.4mm of a6]   (a7)  {\alg{7} \textbf{verified micro-step search} --- beam over one-line edits where the base is reliable};
\node[engv, below=2.4mm of a7]   (a8)  {\alg{8} \textbf{self-authored oracles} $\Rightarrow V_{\mathrm{self}}$ --- dense, grader-free in-loop reward};
\node[engs, below=2.4mm of a8]   (a9)  {\alg{9} \textbf{search-layer controllers} --- compute-stop $\cdot$ QD diversify $\cdot$ step-credit $\cdot$ forgetting gate};
\node[engs, below=2.4mm of a9]   (a10) {\alg{10} \textbf{verified self-repair} --- adopts edit/compose fixes by measured fix-rate ($\to L_2$)};
\node[draw=green!45!black, dashed, rounded corners=3pt, fit=(actor)(a6)(a7)(a8)(a9)(a10), inner sep=2.8mm,
      label={[font=\scriptsize, text=green!45!black, align=center]90:{actor \& verified-search engine (\alg{6}--\alg{10})\\ \emph{generates \& verifies} candidate behaviors}}] (ENG) {};
\node[slowc, right=22mm of actor.east, anchor=north west] (A4) {\alg{4} SGM-CS --- gated harness edits ($L_2$)};
\node[slowc, below=2.2mm of A4] (A5) {\alg{5} SDC-QD --- grows the abstraction library ($L_2$)};
\node[fastc, below=2.2mm of A5] (A3) {\alg{3} PA-COCOA --- counterfactual credit $\to\theta$ ($L_1$)};
\node[fastc, below=2.2mm of A3] (A1) {\alg{1} PPB-CL --- forgetting gate $+$ trust region ($L_1$)};
\node[guardc, below=2.2mm of A1] (A2) {\alg{2} PNMP-A --- anchors the reward model $q$};
\node[draw=black!70, dashed, rounded corners=3pt, fit=(A4)(A5)(A3)(A1)(A2), inner sep=2.6mm,
      label={[font=\scriptsize, align=center]90:{$L_3$ scheduled controllers (\alg{1}--\alg{5})\\ \emph{gate \& select} self-modifications}}] (CTRL) {};
\node[aux, text width=5.2cm, below=20mm of ENG.south] (ENV) {environment $+$ \textbf{held-out grader}\\[-1pt]{\scriptsize terminal measurement only --- never steers the search}};
\node[aux, fill=violet!10, draw=violet!70!black, text width=2.0cm] (RM) at (ENV -| CTRL) {reward\\[-2pt] model $q$};
\node[draw=black!72, fill=yellow!26, rounded corners=2pt, align=center, inner sep=3pt, font=\scriptsize, right=8mm of RM] (LED) {certificate ledger\\[-1pt]{\scriptsize \textsc{accept}/\textsc{hold}/\NSF{} $\cdot$ $\delta$-spend}};
\draw[arr] (POL.east) -- node[lab, above]{deploy $\pol_t$ / act} (ENG.west);
\draw[arr, green!45!black] (ENG.east) -- node[lab, above, align=center]{verified rollouts\\ $+$ process reward $\tilde R$} (CTRL.west);
\draw[arr] (ENG.south) -- node[lab, right]{finalized patch} (ENV.north);
\draw[arr, violet!70!black] (A2.south) to[out=-90,in=90] node[lab, right=1pt, pos=.4]{e-value drift gate} (RM.north);
\draw[arr, violet!70!black, dashed] (RM.west) to[out=180,in=-20] node[lab, below, pos=.55]{$q$ scores rollouts} (ENG.south east);
\draw[arr, black!70] (CTRL.south) to[out=-90,in=0] (LED.east);
\draw[arr, orange!75!black] (CTRL.north) to[out=90,in=90]
   node[lab, pos=.5, above, align=center]{gated self-modifications $\Rightarrow \pol_{t+1}$\\[-1pt] {\scriptsize \alg{3}/\alg{1}$\to L_1$ \quad \alg{4}/\alg{5}$\to L_2$ \quad \alg{2}$\to q$}}
   (POL.north);
\end{tikzpicture}}
\caption{\textbf{The full self-evolution loop (extends Figure~\ref{fig:arch}).} The five scheduled controllers of Figure~\ref{fig:arch} (right, \alg{1}--\alg{5}) can only select among behaviors the frozen policy stack (left, $\pol_t$) already produces. The verifier-tier mechanisms \alg{6}--\alg{10} (centre, green dashed) are the engine that supplies them: the actor and its search operators (\alg{6}/\alg{7}) manufacture diverse candidate patches; self-authored reproduction oracles (\alg{8}) score each candidate with a dense, grader-free reward $V_{\mathrm{self}}$; the search-layer controllers (\alg{9}, $=\alg{1}/\alg{3}/\alg{4}/\alg{5}$ re-aimed) govern the search; and verified self-repair (\alg{10}) fixes the harness's own edit operators. The engine emits \emph{verified} rollouts and process rewards $\tilde R$ that feed the $L_3$ controllers, which gate each self-modification, write the certificate ledger, and close the loop by updating $\pol_t$ (\alg{3}/\alg{1} on $L_1$, \alg{4}/\alg{5} on $L_2$, \alg{2} on the reward model $q$). The held-out grader sits outside the loop, measuring the finalized patch once and never steering it.}
\label{fig:fullloop}
\end{figure}
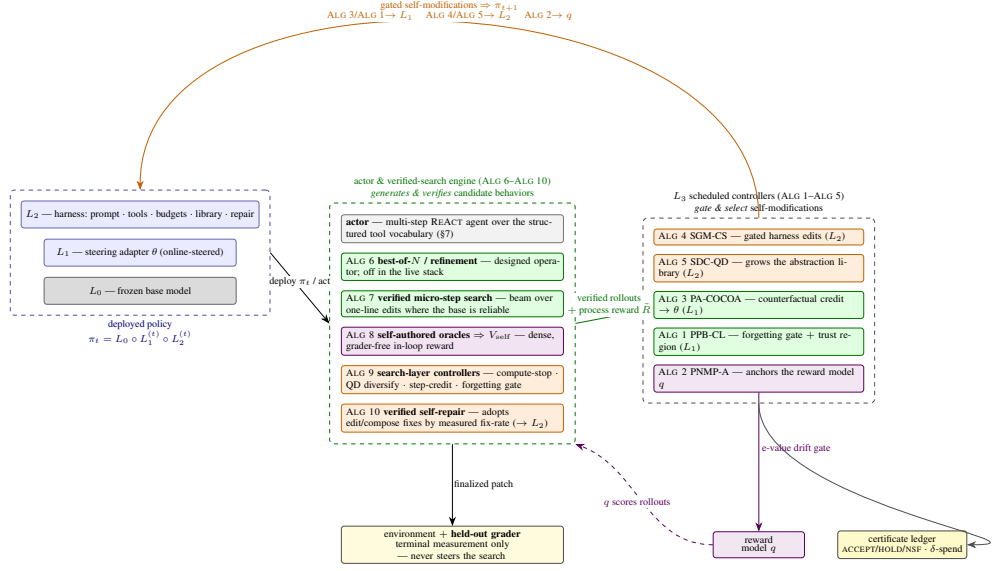

Figure~\ref{fig:arch} shows the four layers and the five scheduled controllers (\alg{1}--\alg{5}), which can only \emph{select} among behaviors the frozen base already produces. Figure~\ref{fig:fullloop} extends it with the verifier-tier mechanisms \alg{6}--\alg{10} (Table~\ref{tab:algmap}) that \emph{generate and verify} those behaviors, supplying the variation and the dense, grader-free signal the controllers consume. The framework is one closed loop over three stages. \textbf{(1)~Policy.} The deployed policy $\pol_t=L_0\circ L_1^{(t)}\circ L_2^{(t)}$ is rolled out. \textbf{(2)~Engine (\alg{6}--\alg{10}).} An actor-and-search engine manufactures diverse candidate patches and scores them against a self-authored verifier: best-of-$N$/refinement varies attempts (\alg{6}), micro-step search relocates the search to one-line edits where even a weak base is reliable (\alg{7}), self-authored reproduction oracles supply the in-loop reward $V_{\mathrm{self}}$ without touching the held-out grader (\alg{8}), the search-layer controllers govern that search (\alg{9}), and verified self-repair fixes the harness's own edit/compose operators (\alg{10}). \textbf{(3)~Controllers (\alg{1}--\alg{5}).} The \emph{verified} rollouts and process rewards the engine emits feed the $L_3$ controllers, which gate every self-modification, write the certificate ledger, and update $\pol_t$ for the next round. The held-out grader sits \emph{outside} the loop: it measures the finalized patch once and never steers the search. The division of labor is the figure's point---\alg{6}--\alg{10} create and verify variation; \alg{1}--\alg{5} gate and select among it; \alg{9} is itself \alg{1}/\alg{3}/\alg{4}/\alg{5} re-aimed onto the search layer (Table~\ref{tab:reaim}).

\section{Certificate Schema}
\label{app:certificate}

Each round of every controller emits an immutable, structured certificate with fields: \code{algorithm}, \code{round}, \code{decision} $\in\{\textsc{accept},\textsc{hold},\textsc{reject},\NSF\}$, \code{delta\_spent} (this round's error spend), \code{cumulative\_delta}, a metrics map, and a free-text note. Controller-specific metrics include: \alg{1} empirical risk, KL to prior, PAC-Bayes penalty and risk bound, forgetting bound, trust radius, gradient norm, damping count, $\varepsilon L$; \alg{2} mean reward, e-value, KL to magnet, slow/fast rates, preference drift; \alg{3} gradient and step norms, path variation, restarts, contribution spread; \alg{4} corrected LCB, per-version value bounds, performative shift, widening flag, baseline-evaluated flag (so the stationarity stream can exclude rounds without a fresh baseline value); \alg{5} solved count, library size, $\Delta J$ (the SWE variant reports it only when the macro committed, alongside a committed-macro flag; the generic controller reports it unconditionally), archive coverage, frontier size, description-length bound. The composite controller merges sub-certificates with per-algorithm metric prefixes into one ledger row per round. The search-layer controllers emit a finer-grained, per-\emph{decision} audit log: an opt-in, decision-neutral JSONL sink writes one flushed row per controller decision (the marginal gain and confidence interval for \alg{4}, the collapsed behavior cells for \alg{5}, the per-branch verified reward for \alg{3}, the measured forgetting for \alg{1}), so a crashed or killed run still leaves a complete, inspectable trail. This per-decision trail is what attributes each flip and regression to a named mechanism in \S\ref{sec:results}.

\section{Default Hyperparameters}
\label{app:hyper}

Defaults in the reference implementation: \alg{1}: $\delta=0.05$, $\tau_{\mathrm{forget}}=0.1$ (raised to $1.1$ above the anytime floor for live multi-step SWE runs), $\varepsilon=0.2$, $L=1$, buffer capacity $m=16$, damping $\rho=0.5$, $\lambda$ set per evaluation to the DV-optimal $\lambda^*=\sqrt{8|\mathcal{A}|A_t}$ ($\lambda=4$ fallback for $A_t\le0$), REINFORCE baseline pooled over a replay window of $8\times$ batch size, learning rate $\eta=0.5$, anchor capacity $A=4$. \alg{2}: $\alpha=0.3$, $\beta=0.1$, $\eta=0.5$, $\tau=0.1$, $\delta_0=0.05$, magnet period $K=5$, slow/fast step exponents $1.0/0.7$ with $a_0=0.5$, preference pairs pooled over the recent replay window. \alg{3}: contribution half-life $h=64$ over a window of $4h$ recent rollouts, drift test from $12$ observations at level $0.01$, exploration floor $\varphi=0.05$. \alg{4}: $\delta_0=0.05$, $\epsilon_{\mathrm{tol}}=0.02$, $\varepsilon=0.1$, CS $\alpha=0.05$ with $\sigma=0.5$ (Hoeffding scale for $[0,1]$ rewards); SWE proposer cycle: force-edit threshold decrements of $2$ down to a floor of $3$ (initialized at half the step budget), step-budget increments of $6$ capped at $36$, guidance rewrites under 60 words; pre-gate pilot of $1$ task with minimum reward $0.05$ (disabled on the first round, when no evidence exists), shadow threshold $1.0$ with retry cap of $2$ per harness; the paired-difference CS runs at scale $\sigma=1.0$. \alg{5}: $\delta=0.05$, solve threshold $1.0$, minimum utility $1.0$; SWE macro-reward bar equal to the corpus bar, downstream lift measured once the contextual pool holds $\ge4$ rewards; generic archive $16\times16$ over (library size, mean program size) in $[0,32]^2$; SWE archive $8\times16$ over (arity, pattern size) in $[0,8]\times[0,16]$, operation windows of length $2$--$3$, corpus bar $0.4$ on the process reward, minimum downstream lift $0.05$ (strictly required; path-literal and single-op macros refused), library capped at $3$ entries per (repository, status) context and $12$ total, retirement over a $24$-rollout window at bad-rate $0.5$. Verifier mechanisms: best-of-$N$ default $n=4$, refinement depth $3$ (best-of-$N$/\alg{6} is removed from the live composite, \S\ref{sec:ablation}; refinement depth $1$ so verify-react stays active; unverified-green cap $0.9$; tie-break judge at score $\ge0.85$ with zero admitted oracles; micro-search issue cap $1.8$k characters; fuzzy-edit hint threshold $0.6$); \code{run\_tests} feedback carries the last $1400$ characters of the failure log. Composite default periods: \alg{1}/\alg{3} every round, \alg{4}/\alg{5} every $2$, \alg{2} every $3$; cost-aware skip after $2$ gainless scheduled rounds.

\section{The SWE Agent Tool Protocol}
\label{app:tools}

The agent replies with exactly one structured action per turn, drawn from the vocabulary

\begin{center}
$\code{list}(d)$\qquad $\code{search}(r)$\qquad $\code{read}(f,\,i{:}j)$\qquad
$\code{edit}(f,\,\textit{old}\!\to\!\textit{new})$\qquad $\code{run\_tests}()$\qquad $\code{submit}()$
\end{center}

over directories $d$, regular expressions $r$, files $f$, line ranges $i{:}j$, and exact text replacements.

Edits require the search text to match exactly once; a miss falls back to a unique whitespace-normalized fuzzy match, and a true miss returns the nearest similar line as a hint (ambiguous matches are refused). Search is regular-expression matching with capped output (at most $60$ matches); read windows report the file's true line total; observations are truncated to a fixed character budget; malformed replies receive a corrective error message and consume a step. \code{run\_tests} is advertised only when a test runner is configured; it applies the current edits, runs the instance's target tests, and returns pass counts plus the tail of the failure log (``submit now'' on a full resolve). When a test runner is available, \code{submit} is blocked until \code{run\_tests} has been run on the latest edits, so the agent cannot submit blind; and an episode that would otherwise end with an empty diff triggers a final no-patch recovery pass (up to two forced-edit attempts at temperature $0$). When the explore$\to$edit budget binds, \code{search}/\code{read}/\code{list} return a corrective message until an edit lands. The final patch is the accumulated edit set, scored by the native evaluator.

\end{document}